\documentclass[lettersize,journal]{IEEEtran}

\usepackage{graphicx}

\usepackage{color}
\usepackage[table]{xcolor}
\usepackage{colortbl}
\usepackage{bm}
\usepackage{amsmath}
\usepackage{epstopdf}

\usepackage{multicol}
\usepackage{multirow}
\usepackage{booktabs}
\usepackage{amssymb}
\usepackage{bbding}
\usepackage{stfloats}

\usepackage{lineno,hyperref}

\usepackage{makecell}
\usepackage{hhline}
\definecolor{linecolor}{rgb}{0.82, 0.94, 0.75}

\usepackage{booktabs} 

\definecolor{greenbg}{rgb}{0.9, 1.0, 0.9} 

\newcommand{\tabincell}[2]{\begin{tabular}{@{}#1@{}}#2\end{tabular}}

\definecolor{lowred}{RGB}{238,18,137}

\definecolor{lowerred}{RGB}{255,110,180}

\newcommand{\dplus}[1]{\fontsize{6pt}{0.1em}\selectfont (\textbf{\textcolor{lowred}{#1}})}

\definecolor{rowunit}{RGB}{128,128,255}
\definecolor{evaunit01green}{RGB}{54,125,189}
\newcommand{\evagreen}[1]{\textcolor{evaunit01green}{#1}}
\newcommand{\dtplus}[1]{\fontsize{6pt}{0.1em}\selectfont (\textbf{\evagreen{#1}})}
\makeatletter
\renewcommand{\maketag@@@}[1]{\hbox{\m@th\normalsize\normalfont#1}}%
\makeatother
\usepackage{diagbox}

\usepackage{amsmath,amsfonts}
\usepackage{algorithmic}
\usepackage{algorithm}
\usepackage{array}
\usepackage[caption=false,font=normalsize,labelfont=sf,textfont=sf]{subfig}
\usepackage{textcomp}
\usepackage{stfloats}
\usepackage{url}
\usepackage{verbatim}
\usepackage{graphicx}
\usepackage{cite}

\begin{document}

\title{Position Anchor Tuning: Towards Efficient Adaptation of Pre-Trained Point Cloud Transformers}

\author{Zheng~Liu,
        Xin~Gao,
        Jinchao~Zhu,
        and~Gao~Huang,~\IEEEmembership{Member,~IEEE}

\thanks{This work has been submitted to the IEEE for possible publication. Copyright may be transferred without notice, after which this version may no longer be accessible.

Zheng~Liu and Xin~Gao are with the School of Automation and Electrical Engineering, University of Science and Technology Beijing, Beijing 100083, China, and also with the Beijing Engineering Research Center of Industrial Spectrum Imaging, Beijing 100083, China.

Jinchao~Zhu is with the College of Software, Nankai University, Tianjin 300350, China.(Corresponding author: jczhu@nankai.edu.cn)

Gao~Huang is with the Department of Automation, BNRist, Tsinghua University, Beijing 100084, China.
}

}

\markboth{Journal of \LaTeX\ Class Files,~Vol.~14, No.~8, August~2026}%
{Shell \MakeLowercase{\textit{et al.}}: A Sample Article Using IEEEtran.cls for IEEE Journals}


\maketitle

\begin{abstract}
Parameter-efficient fine-tuning (PEFT) has recently emerged as a pivotal research direction for adapting pre-trained point cloud transformers to diverse downstream tasks. Although existing methods achieve excellent fine-tuning performance with high parameter efficiency, they ignore inference efficiency. To tackle this problem, a novel PEFT method termed position anchor tuning (PAT) is proposed in this paper. As multi-head attention (MHA) and feed-forward network (FFN) are computation-heavy blocks in pre-trained transformers, PAT decreases their computational cost through token aggregation-expansion pairs. Each pair comprises a token aggregation module (TAM) and a token expansion module (TEM). For MHA and FFN blocks, TAMs extract representative tokens from their input tokens based on position anchors in 3D space. These extracted tokens, rather than the original input tokens, are processed by the blocks, thereby reducing the number of tokens involved in computation. Then, TEMs propagate the learned representations back to the original input tokens. Since TAMs are solely responsible for capturing task-specific representations, base-sharing low-rank adaptation (BSLoRA) is further introduced to enable them to learn such representations effectively with only a small number of trainable parameters. Extensive experiments on widely used benchmarks demonstrate that PAT performs comparably to state-of-the-art methods while incurring significantly lower computational overhead and fewer trainable parameters.

\end{abstract}


\section{Introduction}

\IEEEPARstart{W}{ith} the development of 3D sensor technologies, point cloud analysis has become increasingly important for understanding our 3D world \cite{b1,b2}. Motivated by the remarkable success of transformers in natural language processing (NLP) \cite{b3} and 2D vision tasks \cite{b4}, many researchers have extended transformer architectures to point cloud learning and demonstrated the powerful representation learning abilities of these architectures \cite{b5,b6}. In recent years, pre-training point cloud transformers on large-scale datasets with self-supervised learning has emerged as a standard paradigm to learn general 3D representations \cite{b7,b8}. To boost performance on downstream tasks with limited data, a common practice is to adapt these general representations to task-specific representations by full fine-tuning \cite{b9,b10,b11}. Given a pre-trained backbone, full fine-tuning updates all of its parameters, which incurs high storage costs as different tuned backbones should be stored for different downstream tasks.

Parameter-efficient fine-tuning (PEFT), introducing only a few trainable parameters for each downstream task, is regarded as a promising technique to overcome the limitation of full fine-tuning. Most existing PEFT methods, such as low-rank adaptation (LoRA) \cite{b12,b13}, prompt learning \cite{b14,b15} and adapters \cite{b16,b17}, are proposed for NLP or 2D vision tasks. Unfortunately, due to the inherent properties of point cloud data, it is challenging for these methods to effectively extract informative representations \cite{b18}. PEFT for pre-trained point cloud transformers has recently attracted growing research attention. Related studies primarily focus on designing suitable prompts \cite{b19,b20}, developing learning modules with linear or nonlinear structures \cite{b18,b21,b22}, or exploring strategies for combining the two approaches \cite{b23,b24,b25}. Pre-trained point cloud transformers usually include encoder layers and decoder layers. Existing PEFT methods for point clouds only utilize the general knowledge in pre-trained encoder layers. Specifically, they freeze the encoder layers and employ extra learnable parameters to extract task-relevant representations, which are then fed into the corresponding prediction heads. Although these methods achieve excellent fine-tuning performance with a few trainable parameters, they ignore the inference efficiency on downstream tasks.

All representations in encoder layers are expressed in the form of tokens, which serve as the basic units for computations. The number of tokens involved in computations is a key factor influencing the computational cost, and fewer tokens result in lower cost. Each encoder layer contains a multi-head attention (MHA) block, a feed-forward network (FFN) block and two layer-norm operators. Compared with layer-norm operators, MHA and FFN blocks are far more computationally intensive components. This raises the question: is it possible to transfer the general knowledge embedded in pre-trained MHA and FFN blocks to representative tokens rather than all input tokens, and then directly propagate the transferred knowledge back to the original input tokens?

Point clouds consist of numerous points in 3D space, and spatial neighbors of these points normally share similar feature representations \cite{b26}. Leveraging the spatial information provided by point clouds, we can obtain a set of anchor points and the corresponding local neighbors. Based on these anchor points and their neighbors, it is feasible to extract representative tokens for pre-trained MHA and FFN blocks and subsequently propagate the enriched representations back to the original tokens. This reduces the number of tokens processed by MHA and FFN blocks, thereby lowering computational overhead and improving inference efficiency.

\begin{figure}[htb]
\centering
\includegraphics[trim={5.3cm 7.05cm -1.5cm 5.8cm},clip,width=4.52in]{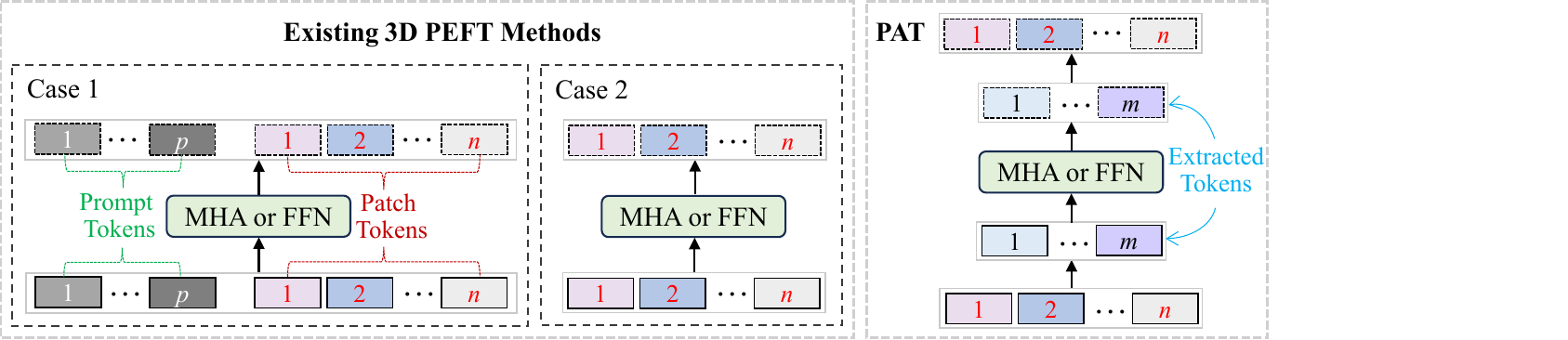}
\caption{Inference differences between PAT and existing 3D PEFT methods in MHA/FFN blocks. $p$, $n$, and $m$ ($m<n$) represent the number of prompt tokens, patch tokens, and extracted tokens, respectively. Existing 3D PEFT methods include two cases: increasing the number of input tokens (Case 1) and keeping the input tokens unchanged (Case 2).}
\label{fig1}
\end{figure}

In this paper, we introduce a novel PEFT method, named position anchor tuning (PAT), to transfer the general knowledge in pre-trained point cloud transformers to downstream tasks. For simplicity, we denote the PEFT methods designed for point clouds as 3D PEFT methods. Fig.~\ref{fig1} illustrates the inference differences between PAT and existing 3D PEFT methods. In MHA and FFN blocks, existing 3D PEFT methods either increase the number of their processed tokens or directly process all patch tokens. In contrast, PAT first extracts representative tokens from these patch tokens based on position anchors in 3D space, then allows the blocks to process the extracted tokens, and finally propagates the obtained information to all the original input tokens. Compared to existing 3D PEFT methods, PAT is more inference-efficient because it reduces the computational cost in MHA and FFN blocks.

PAT inserts token aggregation-expansion pairs, each of which comprises a token aggregation module (TAM) and a token expansion module (TEM), into pre-trained backbones to enable both parameter-efficient and inference-efficient adaptation. Before MHA and FFN blocks, TAMs extract representative tokens from their input tokens. The blocks then process these extracted tokens instead of the original input tokens, thereby reducing computational overhead. TEMs are inserted after the blocks to transfer the obtained representations to the original input tokens. TAMs introduce trainable parameters to capture task-specific representations, whereas TEMs operate in a training-free manner. Accordingly, we propose base-sharing low-rank adaptation (BSLoRA) to enable TAMs to effectively learn such representations with high parameter efficiency. The main contributions are as follows.

\begin{itemize}

 \item We introduce token aggregation-expansion pairs to reduce the number of tokens processed by MHA and FFN blocks, significantly lowering the computational overhead during inference.

 \item BSLoRA incorporates a base-sharing scheme and similarity-based regularization to enhance the learning ability of TAMs, requiring a few trainable parameters and introducing no extra inference cost.

 \item  Extensive experiments show that PAT achieves performance on par with state-of-the-art methods while maintaining advantages in both parameter efficiency and inference efficiency.
\end{itemize}

\section{Related work}

\subsection{PEFT in NLP and 2D Vision}

PEFT has been widely explored in NLP and 2D vision. The mainstream studies can be broadly divided into three classes: prompt tuning, auxiliary-module tuning and hybrid tuning.

\textbf{Prompt tuning} adds task-specific learnable prompts, such as vectors and tokens, to the inputs of pre-trained backbones or their internal modules. For example, Li et al. \cite{b27} prepend continuous prefix vectors to the input tokens; Zhou et al. \cite{b28} use hierarchical visual prompts to encode domain-level and task-specific characteristics; Jiang et al. \cite{b29} boost prompt adaptivity and expressiveness via the dynamic generation of sample-dependent prompts. Among the three classes, \textbf{auxiliary-module tuning}, employing extra modules to adapt the general knowledge in pre-trained backbones, is the most popular one. Auxiliary-module tuning mainly includes side-tuning, adapter-tuning and reparameterization-tuning. In side-tuning, related studies such as UniPT \cite{b30} and MDPD \cite{b31} apply lightweight networks to refine the intermediate features extracted at varying depths. In adapter-tuning, adapters \cite{b16} are typically used to tune the input/output representations of transformer blocks (such as encoder layers \cite{b32}, MHA and FFN \cite{b33}) or learn complementary information for corresponding blocks \cite{b34}. Reparameterization-tuning aims to design learning modules that can be seamlessly integrated into pre-trained models. Accordingly, related methods, such as PiVot \cite{b35}, SSF \cite{b36}, FacT \cite{b37} and PEGO \cite{b38}, introduce no extra inference cost. \textbf{Hybrid tuning} incorporates different PEFT methods into a unified framework. For example, Zhang et al. \cite{b39} explore a unified search space comprising LoRA, VPT, and adapters to select the optimal layer configurations; Zhou et al. \cite{b40} design a search-based approach to automatically configure prefix tokens, serial LoRA and parallel LoRA.

In PAT, BSLoRA, as a variant of LoRA, is employed to learn task-specific representations. BSLoRA introduces only a small number of trainable parameters due to its base-sharing scheme. Recently, LoRA-based methods incorporating sharing schemes have been explored in NLP and 2D vision \cite{b41,b42,b43,b44}. In these methods, both specific information and shared information are entangled, implying that they may adversely influence each other. In contrast, BSLoRA separately learns specific representations and shared representations, and further improves their quality via similarity-based regularization.

\subsection{PEFT in Point Cloud Learning}

Compared to NLP and 2D vision, PEFT for point cloud learning tasks is relatively under-explored. The related studies can be roughly categorized into four branches: grouping-based tuning, interpolation-based tuning, graph-based tuning and locality-agnostic tuning.

\textbf{Grouping-based tuning} regroups the original point patches or utilizes additional group information about point clouds. Zhang et al. \cite{b45} extract new groups from input point clouds as prompt tokens and employ adapters to boost the adaptation ability. Wang et al. \cite{b25} divide point clouds into groups with different scales and tokenize them as prompts. \textbf{Interpolation-based tuning} integrates the interpolations of neighbor tokens into the task-specific representation learning process. MoST \cite{b21} learns the updates of pre-trained weights by structured matrices comprising linear transformations and geometric neighbor interpolations. Ai et al. \cite{b46} propose a geometry-aware prompt method where prompt tokens and original tokens interact with each other in the interpolation-based feature propagation process. In Point-PEFT \cite{b47}, geometry-aware adapters incorporating neighbor interpolations are designed to capture local geometric information. \textbf{Graph-based tuning} represents the relationship among points or point tokens via graph structures. Liang et al. \cite{b18} construct a local graph and a global graph from points, and introduce a lightweight point cloud spectral adapter based on both graphs. IDPT \cite{b19}, PFPT \cite{b20} and GFT \cite{b22} employ EdgeConv \cite{b48}, which operates on layer-wise recomputed graphs, to aggregate local information. Unlike the above three categories, \textbf{locality-agnostic tuning} directly designs adaptation strategies without considering local information. Pointing out that static feature aggregation operators, such as mean or max pooling, limit the fine-tuning performance, Fei et al. \cite{b49} introduce a dynamic aggregation method to replace them. Zhou et al. \cite{b23} adapt the representations in pre-trained backbones using a task-agnostic feature transform strategy, internal prompts and dynamic adapters.

Existing 3D PEFT methods adapt pre-trained point cloud transformers to downstream tasks in a parameter-efficient manner. However, they mainly focus on reducing the number of trainable parameters, ignoring the computational overhead during inference. In contrast, PAT aims to fine-tune pre-trained backbones efficiently, reducing both parameter count and computational cost.

\section{Preliminaries}
\subsection{Point Cloud Transformers}

Due to the powerful representation learning capability of vision transformer (ViT), its structure has been extended to point cloud analysis. A typical point cloud transformer includes three core components: a point tokenizer, encoder layers and a prediction head \cite{b8}.

The point tokenizer partitions each point cloud into local patches by farthest point sampling (FPS) and the $k$-nearest neighbor (KNN) algorithm, and then encodes these patches into tokens to facilitate the subsequent information learning process. Specifically, given a point cloud ${\bf{X}} \in \mathbb{R}^{N \times 3}$ with $N$ points, we can obtain its local patches ${{\bf{X}}_p}$ by
\begin{equation}
{{\bf{X}}_p} = {\rm{KNN}}({\bf{X}},{\bf{P}},k), \label{eq1}
\end{equation}
where ${\bf{P}} = [{{\bf{P}}_1};{{\bf{P}}_2}; \cdots ;{{\bf{P}}_n}] \in \mathbb{R}^{n \times 3}$ represents the $n$ group centers obtained by FPS, and ${{\bf{X}}_p} \in \mathbb{R}^{n \times k \times 3}$ contains the $k$ neighbor points of all group centers. Then, the local patches ${{\bf{X}}_p}$ are encoded into patch tokens ${\bf{E}} \in \mathbb{R}^{n \times d}$ by a lightweight PointNet \cite{b50}, where $d$ is the embedding dimension.  Finally, a class token $cls$ is appended to the patch tokens as the input of the first encoder layer.

Each encoder layer mainly includes an MHA block, an FFN block and two layer-norm (LN) operators. Let ${{\bf{c}}_h} \in \mathbb{R}^{1 \times d}$ and ${{\bf{Z}}_h} \in \mathbb{R}^{n \times d}$ denote the representations of class token $cls$  and patch tokens ${\bf{E}}$ input into the $h$-th ($1 \le h \le L$) encoder layer. Then, we have
\begin{equation}
[{{\bf{\tilde c}}_h};{{\bf{\tilde Z}}_h}] = {\rm{MHA}}({\rm{LN}}([{{\bf{c}}_h};{{\bf{Z}}_h}])) + [{{\bf{c}}_h};{{\bf{Z}}_h}], \label{eq2}
\end{equation}
\begin{equation}
[{{\bf{c}}_{h + 1}};{{\bf{Z}}_{h + 1}}] = {\rm{FFN}}({\rm{LN}}([{{\bf{\tilde c}}_h};{{\bf{\tilde Z}}_h}])) + [{{\bf{\tilde c}}_h};{{\bf{\tilde Z}}_h}], \label{eq3}
\end{equation}
where ${{\bf{\tilde c}}_h}$ and ${{\bf{\tilde Z}}_h}$ are the extracted representations about ${{\bf{c}}_h}$ and ${{\bf{Z}}_h}$. After the $L$-th encoder layer, we can obtain ${{\bf{c}}_{L + 1}}$ and ${{\bf{Z}}_{L + 1}}$, which are fed into the corresponding prediction heads.

\begin{figure*}[!htb]
\centering
\includegraphics[trim={0.8cm 1.17cm -9cm 1cm},clip,width=9.3in]{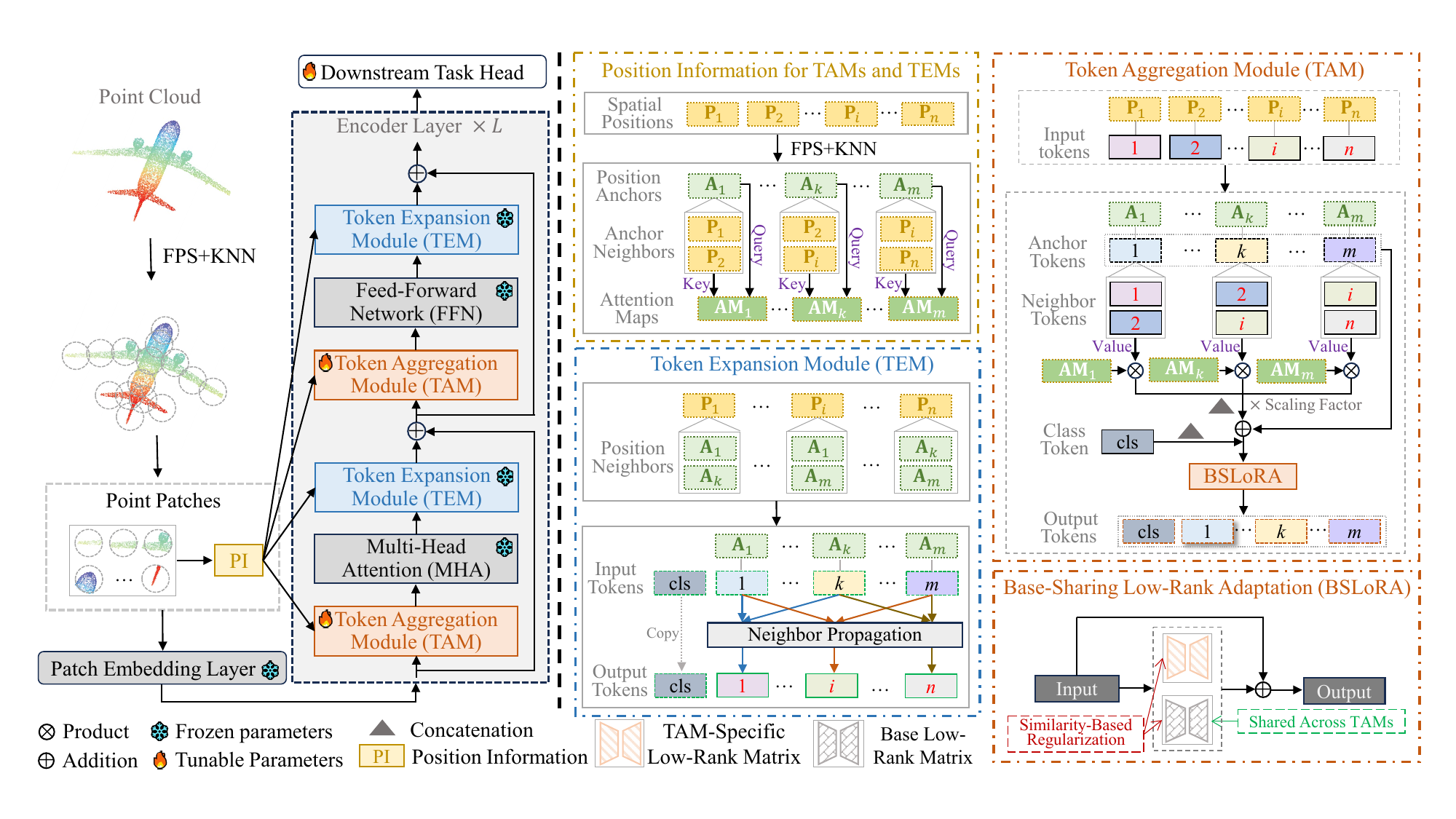}
\caption{Illustration of PAT. For the convenience of presentation, the 3D spatial information required by TAMs and TEMs, including point patch positions $\{ {{\bf{P}}_i}\} _{i = 1}^n$, position anchors $\{ {{\bf{A}}_i}\} _{i = 1}^m$, and the attention maps $\{ {\bf{A}}{{\bf{M}}_i}\} _{i = 1}^m$ between position anchors and their neighbors, is processed within the PI module. TAMs, inserted before MHA or FFN blocks, extract $m$ anchor tokens from $n$ patch tokens based on the provided information, and then apply BSLoRA to effectively adapt these anchor tokens to downstream tasks in a parameter-efficient way.}
\label{fig2}
\end{figure*}

\subsection{Low-Rank Adaptation}

When adapting a pre-trained model to a downstream task, LoRA \cite{b12} focuses on learning the weight updates and assumes that these updates are in low-rank spaces. Given a pre-trained weight ${{\bf{W}}_0} \in \mathbb{R}^{{d_1} \times {d_2}}$, its update $\Delta {{\bf{W}}_0}$ is learned by the product of two low-rank matrices ${{\bf{A}}_0} \in \mathbb{R}^{{d_1} \times v}$ and ${{\bf{B}}_0} \in \mathbb{R}^{v \times {d_2}}$, where $v \ll \min \{ {d_1},{d_2}\} $.  Accordingly, the output ${{\bf{X}}_{{\rm{output}}}}$ can be formulated as
\begin{equation}
{{\bf{X}}_{{\rm{output}}}} = {{\bf{X}}_{{\rm{input}}}}({{\bf{W}}_0} + \Delta {{\bf{W}}_0}) = {{\bf{X}}_{{\rm{input}}}}({{\bf{W}}_0} + {{\bf{A}}_0}{{\bf{B}}_0}),\label{eq4}
\end{equation}
where ${{\bf{X}}_{{\rm{input}}}}$ represents the input.

During the fine-tuning process, ${{\bf{W}}_0}$ is frozen while ${{\bf{A}}_0}$ and ${{\bf{B}}_0}$ are trained on the data of the downstream task. Since $\Delta {{\bf{W}}_0}$ can be incorporated into the pre-trained model by replacing ${{\bf{W}}_0}$ with ${{\bf{W}}_0} + \Delta {{\bf{W}}_0}$, LoRA introduces no additional computational cost at the inference stage.

\section{Position Anchor Tuning}

As shown in Fig.~\ref{fig2}, PAT employs extra modules to learn task-specific representations with both inference efficiency and parameter efficiency. In the backbone, the computations primarily come from MHA and FFN blocks. To improve inference efficiency, token aggregation-expansion pairs, each consisting of a TAM and a TEM, are used to lower the computational cost of the two computation-heavy blocks. A TAM first selects anchor tokens from the input tokens of an MHA (or FFN) block based on the spatial positions of point patches, then incorporates neighbor information into these anchor tokens via spatial-feature attention, and finally feeds the enriched anchor tokens into the block. After that, a TEM propagates the learned representations to the original input tokens by exploiting their spatial similarities to neighboring anchor tokens. In PAT, only TAMs introduce trainable parameters. To reduce the number of trainable parameters while enhancing representation-learning ability, BSLoRA, which incorporates a base-sharing scheme and similarity-based regularization, is further designed for TAMs.

\subsection{Token Aggregation-Expansion Pair}

As illustrated in Section III.A, group centers ${\bf{P}}$ contain the 3D spatial information associated with patch tokens ${\bf{E}}$. In each encoder layer, an MHA block and an FFN block are adopted to extract informative representations for patch tokens and the class token $cls$. For the two blocks, their input and output tokens share the same spatial information existing in ${\bf{P}}$. In 3D space, patch tokens that are spatially close usually have high feature similarity \cite{b26}. This indicates that the number of tokens processed by MHA and FFN blocks can be reduced by extracting a set of spatially representative tokens. Accordingly, we introduce the token aggregation-expansion pair in this subsection.

\noindent\textit{1) Token Aggregation Module (TAM)}

To achieve spatially uniform coverage of the whole 3D space, we select $m$ position anchors ${\bf{A}} = [{{\bf{A}}_1}; \cdots ;{{\bf{A}}_m}] \in \mathbb{R}^{m \times 3}$ by
\begin{equation}
{\bf{A}},{\rm{idx}} = {\rm{FPS}}({\bf{P}}),
\label{eq5}
\end{equation}
where ${\rm{FPS}}( \cdot )$ denotes the FPS algorithm and ${\rm{idx}}$ is the index set of position anchors. For these position anchors, their $u$-nearest neighbors ${\bf{B}} \in \mathbb{R}^{m \times u \times 3}$ from ${\bf{P}}$ are obtained by
\begin{equation}
{\bf{B}},{\rm{IDX}} = {\rm{KNN}}({\bf{P}},{\bf{A}},u),
\label{eq6}
\end{equation}
where ${\rm{IDX}} = [{\rm{ID}}{{\rm{X}}_1}; \cdots ;{\rm{ID}}{{\rm{X}}_m}]$ is the total index set of neighbors and ${\bf{B}}[i] \in \mathbb{R}^{u \times 3}$ represents the $u$ neighbors of ${{\bf{A}}_i}$ with neighbor indices in ${\rm{ID}}{{\rm{X}}_i}$.

In the $h$-th ($1 \le h \le L$) encoder layer, the input tokens for MHA and FFN blocks are $[{\bf{c}}_h^{{\rm{MHA}}};{\bf{Z}}_h^{{\rm{MHA}}}]$ and  $[{\bf{c}}_h^{{\rm{FFN}}};{\bf{Z}}_h^{{\rm{FFN}}}]$, respectively. Since ${\bf{P}}$ provides the 3D coordinates for ${\bf{Z}}_h^{{\rm{MHA}}}$ and ${\bf{Z}}_h^{{\rm{FFN}}}$, we can obtain the corresponding anchor tokens ${\bf{A}}_h^{{\rm{MHA}}} \in \mathbb{R}^{m \times d}$ and ${\bf{A}}_h^{{\rm{FFN}}} \in \mathbb{R}^{m \times d}$ by indexing ${\bf{Z}}_h^{{\rm{MHA}}}$ and ${\bf{Z}}_h^{{\rm{FFN}}}$ with ${\rm{idx}}$. Replacing ${\bf{Z}}_h^{{\rm{MHA}}}$ and ${\bf{Z}}_h^{{\rm{FFN}}}$ with ${\bf{A}}_h^{{\rm{MHA}}}$  and ${\bf{A}}_h^{{\rm{FFN}}}$ directly is a simple way to reduce the involved computations. However, it ignores lots of context information contained in ${\bf{Z}}_h^{{\rm{MHA}}}$ and ${\bf{Z}}_h^{{\rm{FFN}}}$, which negatively affects the quality of learned representations. To solve this problem, the information of corresponding local neighbors is aggregated to enrich ${\bf{A}}_h^{{\rm{MHA}}}$  and ${\bf{A}}_h^{{\rm{FFN}}}$. Suppose ${\bf{B}}_h^{{\rm{MHA}}} \in \mathbb{R}^{m \times u \times d}$ and ${\bf{B}}_h^{{\rm{FFN}}} \in \mathbb{R}^{m \times u \times d}$ are the corresponding neighbor tokens sampled from ${\bf{Z}}_h^{{\rm{MHA}}}$ and ${\bf{Z}}_h^{{\rm{FFN}}}$ with the indices existing in ${\rm{IDX}}$. Based on the pairwise interaction nature of attention, we aggregate the information of neighbor tokens by the spatial-feature attention defined as follows.
\begin{equation}
{\bf{\bar z}}_{h,i}^{{\rm{MHA/FFN}}} = map({{\bf{A}}_i}{{\bf{W}}_a},{\bf{B}}[i]{{\bf{W}}_b}){\bf{B}}_h^{{\rm{MHA/FFN}}}[i],
\label{eq7}
\end{equation}
\begin{equation}
{\bf{\bar Z}}_h^{{\rm{MHA/FFN}}} = [{\bf{\bar z}}_{h,1}^{{\rm{MHA/FFN}}}; \cdots ;{\bf{\bar z}}_{h,m}^{{\rm{MHA/FFN}}}],
\label{eq8}
\end{equation}
where ${\bf{B}}_h^{{\rm{MHA/FFN}}}[i] \in \mathbb{R}^{u \times d}$ represents the neighbor tokens of ${{\bf{A}}_i}$ in the MHA (or FFN) block; $map( \cdot , \cdot )$ denotes the self-attention map; ${\bf{\bar z}}_{h,i}^{{\rm{MHA/FFN}}} \in \mathbb{R}^{1 \times d}$ is the neighbor information aggregated for ${{\bf{A}}_i}$; ${\bf{\bar Z}}_h^{{\rm{MHA/FFN}}} \in \mathbb{R}^{m \times d}$ is the set of aggregated tokens; ${{\bf{W}}_a} \in \mathbb{R}^{3 \times 3},{{\bf{W}}_b} \in \mathbb{R}^{3 \times 3}$ are trainable projection matrices in 3D space. Let ${\bf{\bar A}}_h^{{\rm{MHA}}}$ and ${\bf{\bar A}}_h^{{\rm{FFN}}}$ denote the anchor tokens that are enhanced with local information, we have
\begin{equation}
{\bf{\bar A}}_h^{{\rm{MHA/FFN}}} = {\bf{A}}_h^{{\rm{MHA/FFN}}} + \alpha _h^{{\rm{MHA/FFN}}}{\bf{\bar Z}}_h^{{\rm{MHA/FFN}}},
\label{eq9}
\end{equation}
where $\alpha _h^{{\rm{MHA}}}$ and $\alpha _h^{{\rm{FFN}}}$ are learnable coefficients that modulate the contributions of ${\bf{\bar Z}}_h^{{\rm{MHA}}}$ and ${\bf{\bar Z}}_h^{{\rm{FFN}}}$.

Pre-trained backbones provide general representations that are useful for downstream tasks. Nevertheless, they typically lack task-specific knowledge, which implies that directly feeding $[{\bf{c}}_h^{{\rm{MHA}}};{\bf{\bar A}}_h^{{\rm{MHA}}}]$ and $[{\bf{c}}_h^{{\rm{FFN}}};{\bf{\bar A}}_h^{{\rm{FFN}}}]$ into the MHA and FFN blocks fails to transfer the general knowledge to downstream domains. Inspired by existing studies \cite{b43,b51}, we employ trainable matrices to capture the representation updates for downstream tasks. For convenience, we denote ${\bf{\bar X}}_h^{{\rm{MHA}}}$ and ${\bf{\bar X}}_h^{{\rm{FFN}}}$ as the adapted tokens to be processed by the two blocks. We formulate them as
\begin{equation}
{\bf{\bar X}}_h^{{\rm{MHA}}} = [{\bf{c}}_h^{{\rm{MHA}}};{\bf{\bar A}}_h^{{\rm{MHA}}}] + [{\bf{c}}_h^{{\rm{MHA}}};{\bf{\bar A}}_h^{{\rm{MHA}}}]{\bf{W}}_h^{{\rm{MHA}}}.
\label{eq10}
\end{equation}
\begin{equation}
{\bf{\bar X}}_h^{{\rm{FFN}}} = [{\bf{c}}_h^{{\rm{FFN}}};{\bf{\bar A}}_h^{{\rm{FFN}}}] + [{\bf{c}}_h^{{\rm{FFN}}};{\bf{\bar A}}_h^{{\rm{FFN}}}]{\bf{W}}_h^{{\rm{FFN}}},
\label{eq11}
\end{equation}
where $[{\bf{c}}_h^{{\rm{MHA}}};{\bf{\bar A}}_h^{{\rm{MHA}}}]{\bf{W}}_h^{{\rm{MHA}}}$ and $[{\bf{c}}_h^{{\rm{FFN}}};{\bf{\bar A}}_h^{{\rm{FFN}}}]{\bf{W}}_h^{{\rm{FFN}}}$ are the representation updates learned by trainable matrices ${\bf{W}}_h^{{\rm{MHA}}} \in \mathbb{R}^{d \times d}$ and ${\bf{W}}_h^{{\rm{FFN}}} \in \mathbb{R}^{d \times d}$. After the MHA and FFN blocks, we have
\begin{equation}
[{\bf{\hat c}}_h^{{\rm{MHA}}};{\bf{\hat Z}}_h^{{\rm{MHA}}}] = {\rm{MHA}}({\bf{\bar X}}_h^{{\rm{MHA}}}),
\label{eq12}
\end{equation}
\begin{equation}
[{\bf{\hat c}}_h^{{\rm{FFN}}};{\bf{\hat Z}}_h^{{\rm{FFN}}}] = {\rm{FFN}}({\bf{\bar X}}_h^{{\rm{FFN}}}),
\label{eq13}
\end{equation}
where ${\bf{\hat Z}}_h^{{\rm{MHA/FFN}}}$ and ${\bf{\hat c}}_h^{{\rm{MHA/FFN}}}$ are the transformed representations about patch tokens and the class token.

\noindent\textit{2) Token Expansion Module (TEM)}

After obtaining $[{\bf{\hat c}}_h^{{\rm{MHA}}};{\bf{\hat Z}}_h^{{\rm{MHA}}}]$ and $[{\bf{\hat c}}_h^{{\rm{FFN}}};{\bf{\hat Z}}_h^{{\rm{FFN}}}]$ in the $h$-th encoder layer, we expect to  propagate the learned information in ${\bf{\hat Z}}_h^{{\rm{MHA}}}$ and ${\bf{\hat Z}}_h^{{\rm{FFN}}}$ to ${\bf{Z}}_h^{{\rm{MHA}}}$ and ${\bf{Z}}_h^{{\rm{FFN}}}$, respectively. Both ${\bf{\hat Z}}_h^{{\rm{MHA}}}$ and ${\bf{\hat Z}}_h^{{\rm{FFN}}}$ are feature representations for position anchors ${\bf{A}}$. ${\bf{Z}}_h^{{\rm{MHA}}}$ and ${\bf{Z}}_h^{{\rm{FFN}}}$ are the extracted representations for patch tokens and both of them share the same spatial information provided by ${\bf{P}}$. Since spatial neighbors usually have similar feature representations \cite{b26}, we utilize the neighbor similarities between ${\bf{P}}$ and ${\bf{A}}$ to obtain the representations for ${\bf{Z}}_h^{{\rm{MHA}}}$ and ${\bf{Z}}_h^{{\rm{FFN}}}$. Following PointNet++ \cite{b52}, we use the inverse distance as the pairwise similarity. Assuming ${\rm{i}}{{\rm{d}}_i}$ stores the indices of the $u$-nearest neighbors of ${{\bf{P}}_i}$  with respect to ${\bf{A}}$, we have
\begin{equation}
{\bf{\tilde z}}_{h,i}^{{\rm{MHA/FFN}}} = \sum\limits_{j \in {\rm{i}}{{\rm{d}}_i}} {\frac{{di{s^{ - 1}}({{\bf{P}}_i},{{\bf{A}}_j})}}{{{t_i}}}} {\bf{\hat Z}}_{h,j}^{{\rm{MHA/FFN}}},
\label{eq14}
\end{equation}
\begin{equation}
{\bf{\tilde Z}}_h^{{\rm{MHA/FFN}}} = [{\bf{\tilde z}}_{h,1}^{{\rm{MHA/FFN}}}; \cdots ;{\bf{\tilde z}}_{h,n}^{{\rm{MHA/FFN}}}],
\label{eq15}
\end{equation}
where ${\bf{\tilde Z}}_h^{{\rm{MHA/FFN}}} \in \mathbb{R}^{n \times d}$ represents the propagated feature tokens, $dis( \cdot , \cdot )$ denotes the squared Euclidean distance between two vectors, ${t_i} = \sum\limits_{j \in {\rm{i}}{{\rm{d}}_i}} {di{s^{ - 1}}({{\bf{P}}_i},{{\bf{A}}_j})} $, ${\bf{\hat Z}}_{h,j}^{{\rm{MHA/FFN}}} \in \mathbb{R}^{1 \times d}$ is the $j$-th feature token in ${\bf{\hat Z}}_h^{{\rm{MHA/FFN}}}$. As for the class token representations ${\bf{c}}_h^{{\rm{MHA}}}$ and ${\bf{c}}_h^{{\rm{FFN}}}$, they just pass through TEMs without any processing.

\subsection{Base-Sharing Low-Rank Adaptation}

As shown in equations (10) and (11), TAMs apply trainable matrices $\{ {\bf{W}}_h^{{\rm{MHA}}}\} _{h = 1}^L$ and $\{ {\bf{W}}_h^{{\rm{FFN}}}\} _{h = 1}^L$ to capture task-specific representations. For simplicity, we denote the $2L$ projection matrices as $\{ {{\bf{R}}_i}\} _{i = 1}^{2L}$  where $\{ {{\bf{R}}_i}\} _{i = 1}^L = \{ {\bf{W}}_h^{{\rm{MHA}}}\} _{h = 1}^L$ and $\{ {{\bf{R}}_i}\} _{i = L + 1}^{2L} = \{ {\bf{W}}_h^{{\rm{FFN}}}\} _{h = 1}^L$. These projection matrices introduce $2L{d^2}$ trainable parameters, which is extremely parameter-inefficient during fine-tuning. Let $\{ {\bf{X}}_i^{{\rm{input}}}\} _{i = 1}^L = \{ [{\bf{c}}_h^{{\rm{MHA}}};{\bf{\bar A}}_h^{{\rm{MHA}}}]\} _{h = 1}^L$, $\{ {\bf{X}}_i^{{\rm{input}}}\} _{i = L + 1}^{2L} = \{ [{\bf{c}}_h^{{\rm{FFN}}};{\bf{\bar A}}_h^{{\rm{FFN}}}]\} _{h = 1}^L$, $\{ {\bf{X}}_i^{{\rm{output}}}\} _{i = 1}^L = \{ {\bf{\bar X}}_h^{{\rm{MHA}}}\} _{h = 1}^L$ and $\{ {\bf{X}}_i^{{\rm{output}}}\} _{i = L + 1}^{2L} = \{ {\bf{\bar X}}_h^{{\rm{FFN}}}\} _{h = 1}^L$, equations (10) and (11) can be written as the following unified form.
\begin{equation}
{\bf{X}}_i^{{\rm{output}}} = {\bf{X}}_i^{{\rm{input}}} + {\bf{X}}_i^{{\rm{input}}}{{\bf{R}}_i} = {\bf{X}}_i^{{\rm{input}}}({\bf{I}} + {{\bf{R}}_i}),
\label{eq16}
\end{equation}
where ${\bf{I}}$ represents the identity matrix with proper size.

Equation (16) is a special case of LoRA (see equation (4)) when ${\bf{I}}$ is viewed as the given pre-trained weight with ${{\bf{R}}_i}$ being its weight update. Since the representation update ${\bf{X}}_i^{{\rm{input}}}{{\bf{R}}_i}$ solely relies on the weight update ${{\bf{R}}_i}$, we focus on the learning process of ${{\bf{R}}_i}$. According to the theory of LoRA \cite{b12}, the updates of pre-trained weights lie in low-rank spaces, meaning that the weight updates are low-rank matrices. In LoRA, each of these low-rank matrices is learned by the low-rank module (LRM), which includes a down-projection matrix and an up-projection matrix. Accordingly, we reformulate equation (16) as
\begin{equation}
{\bf{X}}_i^{{\rm{output}}} = {\bf{X}}_i^{{\rm{input}}} + {\bf{X}}_i^{{\rm{input}}}{{\bf{R}}_i} = {\bf{X}}_i^{{\rm{input}}} + {\bf{X}}_i^{{\rm{input}}}{{\bf{A}}_i}{{\bf{B}}_i},
\label{eq17}
\end{equation}
where $r \ll d$, ${{\bf{A}}_i} \in \mathbb{R}^{d \times r}$ and ${{\bf{B}}_i} \in \mathbb{R}^{r \times d}$ represent the down-projection and up-projection matrices for ${{\bf{R}}_i}$. LoRA is a popular PEFT method widely discussed in NLP and 2D vision. By learning low-rank matrix ${{\bf{R}}_i}$ with ${{\bf{A}}_i}{{\bf{B}}_i}$, the trainable parameter count is $4Ldr$, which is generally much smaller than $2L{d^2}$. With $rank( \cdot )$ representing the rank function, we have $rank({{\bf{R}}_i}) = rank({{\bf{A}}_i}{{\bf{B}}_i}) \le r$, meaning that $r$ determines the rank upper bound of ${{\bf{R}}_i}$. In practice, a small $r$ usually limits the learning capacity of low-rank matrices and thus results in poor fine-tuning performance \cite{b17}, while a large $r$ significantly increases the number of trainable parameters, which may lead to overfitting due to parameter redundancy \cite{b41}.

Currently, some sharing schemes \cite{b41,b42,b43,b44} have been proposed to reduce parameter redundancy when $r$ is large. In these schemes, down/up-projection matrices share some or even all components. These methods can be readily extended to learn the low-rank matrices $\{ {{\bf{A}}_i}{{\bf{B}}_i}\} _{i = 1}^{2L}$. For convenience, we denote the shared down-projection and up-projection components as ${{\bf{A}}_{{\rm{shared}}}}$ and ${{\bf{B}}_{{\rm{shared}}}}$. Projection matrices ${{\bf{A}}_i}$ and ${{\bf{B}}_i}$ are then formulated as  $f({{\bf{A}}_{{\rm{shared}}}},{{\bf{\bar A}}_i})$ and $g({{\bf{B}}_{{\rm{shared}}}},{{\bf{\bar B}}_i})$, where ${{\bf{A}}_i}$(${{\bf{B}}_i}$) is composed of ${{\bf{A}}_{{\rm{shared}}}}$ and ${{\bf{\bar A}}_i}$ (${{\bf{B}}_{{\rm{shared}}}}$ and ${{\bf{\bar B}}_i}$); ${{\bf{\bar A}}_i}$ and ${{\bf{\bar B}}_i}$ are TAM-specific matrices;  $f( \cdot , \cdot )$ and $g( \cdot , \cdot )$ represent the corresponding compositional relationships. For example, in VeRA \cite{b42}, ${{\bf{A}}_i} = {{\bf{A}}_{{\rm{shared}}}}{{\bf{\bar A}}_i}$ and ${{\bf{B}}_i} = {{\bf{B}}_{{\rm{shared}}}}{{\bf{\bar B}}_i}$. Within these sharing schemes, ${{\bf{A}}_{{\rm{shared}}}}$ and ${{\bf{B}}_{{\rm{shared}}}}$ are employed to learn the common knowledge across TAMs while $\{ {{\bf{\bar A}}_i}\} _{i = 1}^{2L}$ and $\{ {{\bf{\bar B}}_i}\} _{i = 1}^{2L}$ are applied to capture TAM-specific knowledge, where common knowledge is either randomly initialized or learned jointly with TAM-specific knowledge. Since the two kinds of knowledge are entangled, TAM-specific components may struggle to capture representative information that relies on randomly initialized common knowledge, or may interfere with the learning processes of shared components.

\begin{figure}[!htb]
\centering
\includegraphics[trim={0.1cm 0.50cm 0.0cm 0.36cm},clip,width=3.5in]{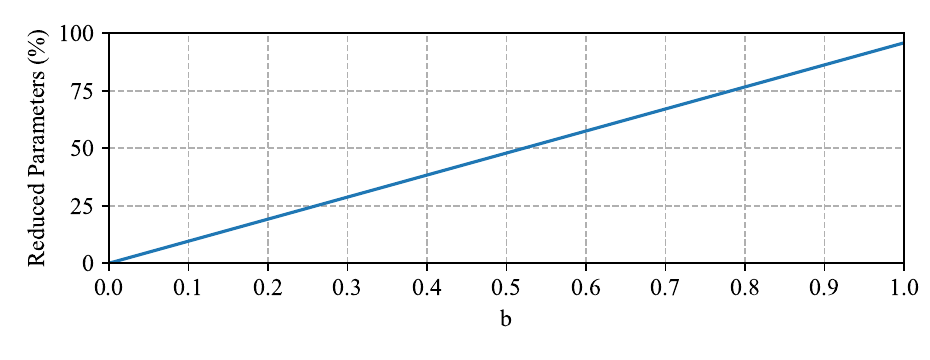}
\caption{Parameter reduction (\%) achieved by BSLoRA compared to LoRA for varying values of $b$.}
\label{fig3}
\end{figure}

To address the above issues, a novel LoRA-based method named BSLoRA is introduced in this subsection. Specifically, we divide each LRM into two independent parts. One is used to learn common knowledge across TAMs and the other is adopted to extract TAM-specific knowledge. In BSLoRA, ${{\bf{R}}_i}$ is formulated as
\begin{equation}
{{\bf{R}}_i} = {\beta _i}{{\bf{A}}_{base}}{{\bf{B}}_{base}} + {\gamma _i}{{\bf{D}}_i}{{\bf{U}}_i},
\label{eq18}
\end{equation}
where ${{\bf{A}}_{base}} \in \mathbb{R}^{d \times br}$ and ${{\bf{B}}_{base}} \in \mathbb{R}^{br \times d}$ are base down-projection and up-projection matrices shared across TAMs, ${{\bf{D}}_i} \in \mathbb{R}^{d \times (1 - b)r}$ and ${{\bf{U}}_i} \in \mathbb{R}^{(1 - b)r \times d}$ are TAM-specific down-projection and up-projection matrices, $0 < b < 1$, ${\beta _i}$ and ${\gamma _i}$ are trainable coefficients that weight the contributions of ${{\bf{A}}_{base}}{{\bf{B}}_{base}}$ and ${{\bf{D}}_i}{{\bf{U}}_i}$. According to equation (18), we have $rank({{\bf{R}}_i}) \le rank({\beta _i}{{\bf{A}}_{base}}{{\bf{B}}_{base}}) + rank({\gamma _i}{{\bf{D}}_i}{{\bf{U}}_i}) \le r$, indicating that BSLoRA maintains the same rank upper bound as LoRA. For $\{ {{\bf{R}}_i}\} _{i = 1}^{2L}$, BSLoRA introduces $2dbr + 4Ld(1 - b)r$ trainable parameters. Because $4Ldr > 2dbr + 4Ld(1 - b)r$, BSLoRA is more parameter-efficient than LoRA (see equation (17)). Taking the pre-trained Point-MAE backbone \cite{b8} as an example, we show the percentage of parameter reduction achieved by BSLoRA compared to LoRA in Fig.~\ref{fig3}. The comparison results illustrate that BSLoRA has a significant advantage over LoRA in parameter compression for large values of $b$. Even with $b = 0.1$, BSLoRA reduces the parameter count by approximately 10\%.

For convenience, we denote the tokens processed by ${{\bf{R}}_i}$ as ${\bf{X}}_i^{{\rm{input}}} = ({{\bf{f}}_{i,1}};{{\bf{f}}_{i,2}}; \cdots ;{{\bf{f}}_{i,m + 1}})$ where ${{\bf{f}}_{i,j}}$ represents the $j$-th token. After ${{\bf{f}}_{i,j}}$ is transformed by ${{\bf{R}}_i}$, we can obtain its base representation ${\bf{f}}_{i,j}^{{\rm{base}}} = {{\bf{f}}_{i,j}}{{\bf{A}}_{base}}{{\bf{B}}_{base}}$ and its TAM-specific representation  ${\bf{f}}_{i,j}^{{\rm{specific}}} = {{\bf{f}}_{i,j}}{{\bf{D}}_i}{{\bf{U}}_i}$. The shared component ${{\bf{A}}_{base}}{{\bf{B}}_{base}}$ extracts common knowledge across TAMs. Since TAMs are inserted at different positions of the backbone, they generally require $\{ {{\bf{R}}_i}\} _{i = 1}^{2L}$ to produce different outputs, implying that $\{ {{\bf{D}}_i}{{\bf{U}}_i}\} _{i = 1}^{2L}$ should extract different representations for these inserted modules. Correspondingly, within each ${{\bf{R}}_i}$, the base representation is expected to differ from the TAM-specific representation. For each ${{\bf{f}}_{i,j}}$, the two representations are vectors of the same dimension. We adopt the widely used cosine similarity ${s_{ij}}$ to measure the relationship between ${\bf{f}}_{i,j}^{{\rm{base}}}$ and ${\bf{f}}_{i,j}^{{\rm{specific}}}$ as follows:
\begin{equation}
{s_{ij}} = \frac{{{\bf{f}}_{i,j}^{{\rm{base}}}{{({\bf{f}}_{i,j}^{{\rm{specific}}})}^{\rm{T}}}}}{{\left| {{\bf{f}}_{i,j}^{{\rm{base}}}} \right|\left| {{\bf{f}}_{i,j}^{{\rm{specific}}}} \right|}},
\label{eq19}
\end{equation}
where ${( \cdot )^{\rm{T}}}$ is the transpose operator and $\left|  \cdot  \right|$ denotes the vector's magnitude. To enlarge the difference between base representations and TAM-specific representations, we regularize them by minimizing the following similarity-regularization term $R$.
\begin{equation}
R = \frac{1}{{2L(m + 1)}}\sum\limits_{i = 1}^{2L} {\sum\limits_{j = 1}^{m + 1} {{{({s_{ij}})}^2}} } .
\label{eq20}
\end{equation}

\subsection{Objective Function}
Token aggregation-expansion pairs introduce no extra loss term. BSLoRA divides the projection matrices into base parts and TAM-specific parts, and the difference between them is regularized by $R$.  Since $R$ is sample-dependent, we can obtain $bs$ values of $R$ with $bs$ batch-training samples. Here, the average value ${\rm{avg(}}R{\rm{)}}$ is taken as the final regularization term. Therefore, the objective function $o$ of PAT is formulated as
\begin{equation}
o = los{s_{task}} + \eta {\rm{avg}}(R),
\label{eq21}
\end{equation}
where $\eta $ is a balance parameter and $los{s_{task}}$ represents the task-specific loss between predictions and ground-truth labels.

\subsection{Analysis of Inference Computation}
All TAMs share the same position anchors and all TEMs share the same spatial similarities. This implies that some operators such as the attention maps and neighbor distances can be calculated only once. Here, we first discuss the computations incurred by TAMs and TEMs, and then analyze the reduction in inference cost.

The main computations in TAMs involve learning task-specific representations via projection matrices $\{ {\bf{W}}_h^{{\rm{MHA}}}\} _{h = 1}^L$ and $\{ {\bf{W}}_h^{{\rm{FFN}}}\} _{h = 1}^L$. Fortunately, these projection matrices can be seamlessly incorporated into the corresponding MHA and FFN blocks, thus avoiding the extra computational cost at the inference stage. Without loss of generality, we take the $h$-th encoder layer as an example. In the MHA block, the query, key and value matrices are calculated through distinct projection weights. Given the input tokens $[{\bf{c}}_h^{{\rm{MHA}}};{\bf{\bar A}}_h^{{\rm{MHA}}}]$ (see equation (10)), we have
\begin{equation}
{\bf{X}}_h^{Q/K/V} = [{\bf{c}}_h^{{\rm{MHA}}};{\bf{\bar A}}_h^{{\rm{MHA}}}]({\bf{I}} + {\bf{W}}_h^{{\rm{MHA}}}){\bf{W}}_h^{Q/K/V},
\label{eq22}
\end{equation}
where ${\bf{X}}_h^{Q/K/V}$ represents the query/key/value matrix and ${\bf{W}}_h^{Q/K/V}$ is the corresponding projection weight. Based on equation (11), in the FFN block, we have
\begin{equation}
{\bf{X}}_h^{{\rm{FFN}},1} = [{\bf{c}}_h^{{\rm{FFN}}};{\bf{\bar A}}_h^{{\rm{FFN}}}]({\bf{I}} + {\bf{W}}_h^{{\rm{FFN}}}){\bf{W}}_h^{{\rm{FFN}},1},
\label{eq23}
\end{equation}
where ${\bf{W}}_h^{{\rm{FFN}},1}$ denotes the first projection matrix in the FFN block and ${\bf{X}}_h^{{\rm{FFN}},1}$ represents the obtained representations. By updating ${\bf{W}}_h^{Q/K/V}$ and ${\bf{W}}_h^{{\rm{FFN}},1}$ with $({\bf{I}} + {\bf{W}}_h^{{\rm{MHA}}}){\bf{W}}_h^{Q/K/V}$ and $({\bf{I}} + {\bf{W}}_h^{{\rm{FFN}}}){\bf{W}}_h^{{\rm{FFN}},1}$  during inference, we can save the main computations brought by TAMs. Therefore, the projection matrices in TAMs introduce no additional computational overhead. As shown in equations (14) and (15), TEMs incur little computational cost.

\begin{table*}[tp]

  \centering
  \renewcommand{\arraystretch}{0.94}
  \setlength\tabcolsep{3.1pt}
\footnotesize
  \caption{Classification accuracy (\%) on the ModelNet40 dataset and three variants of the ScanObjectNN dataset. \#Param. (M) denotes the average number of trainable parameters in millions. \#GFLOPs denotes the average inference GFLOPs of backbones. Among 3D PEFT methods, the best results are in bold and the second-best results are underlined.}
    \label{tab1}
    \begin{tabular}{lccccccccc}
    \toprule
    \multirow{2}{*}{Pre-trained model} &\multirow{2}{*}{Fine-tuning strategy} &\multirow{2}{*}{Reference} &\multirow{2}{*}{\#Param. (M)} &\multirow{2}{*}{\#GFLOPs} &\multicolumn{3}{c}{ScanObjectNN} &\multirow{2}{*}{ModelNet40}\\
    \cmidrule(lr){6-8}
    & & & & &OBJ\_BG & OBJ\_ONLY &PB\_T50\_RS     \\
    \hline
   \multirow{9}{*}{\tabincell{c}{Point-MAE~\cite{b8}}} & \textcolor{gray}{Full}  & & \textcolor{gray}{22.1 (100\%)}  & \textcolor{gray}{4.76(-)} & \textcolor{gray}{90.02} & \textcolor{gray}{88.29} & \textcolor{gray}{85.18} & \textcolor{gray}{93.2}\\
    & IDPT~\cite{b19} &ICCV 23  & 1.7 (7.69\%) & 7.1\dtplus{$\uparrow$49.16\%} &\underline{91.22}\dplus{+1.20} &\underline{90.02}\dplus{+1.73} & 84.94{\dtplus{-0.24}}  &93.3\dplus{+0.1}\\
    & DAPT~\cite{b23} &CVPR 24& 1.1 (4.97\%)  & 5.0\dtplus{$\uparrow$5.04\%} & 90.88\dplus{+0.86} &\textbf{90.19}\dplus{+1.90} & 85.08{\dtplus{-0.10}}  & \underline{93.5}{\dplus{+0.3}}  \\
    & Point-PEFT~\cite{b47} &AAAI 24 &0.7 (3.13\%)  & 7.6\dtplus{$\uparrow$37.37\%} & 89.67{\dtplus{-0.35}} & 88.98\dplus{+0.69} & 84.91{\dtplus{-0.27}} &93.3\dplus{+0.1} \\

    & PointGST~\cite{b18}  &TPAMI 25&\underline{0.6} (2.77\%) &\underline{4.8}\dtplus{$\uparrow$0.84\%} &\textbf{91.74}\dplus{+1.72} &\textbf{90.19}\dplus{+1.90} &85.29\dplus{+0.11}  & \underline{93.5}{\dplus{+0.3}} \\

    & PointLoRA~\cite{b25} &CVPR 25 & 0.8 (3.43\%)  &5.1\dtplus{$\uparrow$7.14\%} & 90.71\dplus{+0.69} &89.33\dplus{+1.04} &\textbf{85.53}\dplus{+0.35}  & 93.3{\dplus{+0.1}} \\
    & PPT~\cite{b45} &ACMMM 25 & 1.1 (3.43\%)  &11.3\dtplus{$\uparrow$137.4\%} & 91.05\dplus{+1.03} &89.67\dplus{+1.38} &85.01\dtplus{-0.17}  & \textbf{93.6}{\dplus{+0.4}} \\
    & GFT~\cite{b22} &WACV 26 & 0.7 (3.43\%)  &6.4\dtplus{$\uparrow$34.45\%} & 90.88\dplus{+0.86} &89.33\dplus{+1.04} &85.05\dtplus{-0.13}  & 93.2{\dplus{+0.0}} \\

    & PAT(\textbf{ours})&This paper & \textbf{0.5} (2.26\%) & \textbf{4.2}\dplus{$\downarrow$11.76\%} &91.05\dplus{+1.03} &\underline{90.02}\dplus{+1.73} &\underline{85.46}\dplus{+0.28}  & \underline{93.5}{\dplus{+0.3}} \\
    \hline

    \multirow{9}{*}{\tabincell{c}{RECON~\cite{b9}}} & \textcolor{gray}{Full}  & & \textcolor{gray}{22.1 (100\%)}  & \textcolor{gray}{4.76} & \textcolor{gray}{94.32} & \textcolor{gray}{92.77} & \textcolor{gray}{90.01} & \textcolor{gray}{92.5}\\
    & IDPT~\cite{b19} &ICCV 23  & 1.7 (7.69\%) & 7.1\dtplus{$\uparrow$49.16\%} & 93.29{\dtplus{-1.03}} & 91.57{\dtplus{-1.20}} & 87.27{\dtplus{-2.74}}   & 93.4\dplus{+0.9}\\
    & DAPT~\cite{b23} &CVPR 24& 1.1 (4.97\%)  & 5.0\dtplus{$\uparrow$5.04\%} &94.32\dplus{+0.00} & 92.43{\dtplus{-0.34}} &89.38{\dtplus{-0.63}}  & \underline{93.5}{\dplus{+1.0}}  \\
    & Point-PEFT~\cite{b47} &AAAI 24 & 0.7 (3.13\%)  & 7.6\dtplus{$\uparrow$37.37\%} & 91.91{\dtplus{-2.41}} & 90.19{\dtplus{-2.58}} & 86.36{\dtplus{-3.65}} &93.3\dplus{+0.8} \\

    & PointGST~\cite{b18}  &TPAMI 25&\underline{0.6} (2.77\%)  &\underline{4.8}\dtplus{$\uparrow$0.84\%} &\underline{94.49}\dplus{+0.17} &\underline{92.94}\dplus{+0.17} & \textbf{89.49}{\dtplus{-0.52}}  &\textbf{93.6}{\dplus{+1.1}} \\

    & PointLoRA~\cite{b25} &CVPR 25 & 0.8 (3.43\%)  &5.1\dtplus{$\uparrow$7.14\%}  &93.46{\dtplus{-0.86}}  &91.22{\dtplus{-1.55}}  &88.65{\dtplus{-1.36}}   &93.4\dplus{+0.9}  \\
    &PPT~\cite{b45} &ACMMM 25 & 1.1 (3.43\%)  &11.3\dtplus{$\uparrow$137.4\%}  &94.15{\dtplus{-0.17}}  &\textbf{93.12}{\dplus{+0.35}}  &89.24{\dtplus{-0.77}}   &\textbf{93.6}\dplus{+1.1}  \\
    &GFT~\cite{b22} &WACV 26 & 0.7 (3.43\%)  &6.4\dtplus{$\uparrow$34.45\%}  &93.80{\dtplus{-0.52}}  &90.88{\dtplus{-1.89}}  &88.55{\dtplus{-1.46}}   &93.3\dplus{+0.8}  \\
    &PAT(\textbf{ours}) &This paper & \textbf{0.5} (2.26\%)  & \textbf{4.2}\dplus{$\downarrow$11.76\%} & \textbf{94.84}\dplus{+0.52} &\underline{92.94}\dplus{+0.17} & \underline{89.45}{\dtplus{-0.56}}  & \textbf{93.6}{\dplus{+1.1}} \\

    \hline

    \multirow{9}{*}{\tabincell{c}{Point-BERT~\cite{b10}}} & \textcolor{gray}{Full}  & & \textcolor{gray}{22.1 (100\%)}  & \textcolor{gray}{4.76} & \textcolor{gray}{87.43} & \textcolor{gray}{88.12} & \textcolor{gray}{83.07} & \textcolor{gray}{92.7}\\
    & IDPT~\cite{b19} &ICCV 23  & 1.7 (7.69\%) & 7.1\dtplus{$\uparrow$49.16\%} & 88.12\dplus{+0.69} & 88.30\dplus{+0.18} & 83.69\dplus{+0.62}   & 92.6{\dtplus{-0.1}}\\
    & DAPT~\cite{b23} &CVPR 24& 1.1 (4.97\%)  & 5.0\dtplus{$\uparrow$5.04\%} & 91.05\dplus{+3.62} &\textbf{89.67}\dplus{+1.55} &\underline{85.43}\dplus{+2.36}  & 93.1{\dplus{+0.4}}  \\
    & Point-PEFT~\cite{b47} &AAAI 24 & 0.7 (3.13\%)  & 7.6\dtplus{$\uparrow$37.37\%} & 88.81\dplus{+1.38} &\textbf{89.67}\dplus{+1.55} & 85.00\dplus{+1.93} &\underline{93.4}\dplus{+0.7} \\

    & PointGST~\cite{b18}  &TPAMI 25&\underline{0.6} (2.77\%)  &\underline{4.8}\dtplus{$\uparrow$0.84\%} &\textbf{91.39}\dplus{+3.96} &\textbf{89.67}\dplus{+1.55} & \textbf{85.64}\dplus{+2.57}  & \underline{93.4}{\dplus{+0.7}} \\

    & PointLoRA~\cite{b25} &CVPR 25 & 0.9 (4.07\%)  & 5.1\dtplus{$\uparrow$7.14\%} & 89.85\dplus{+2.42} &\underline{88.98}\dplus{+0.86} &84.63\dplus{+1.56}  &93.2{\dplus{+0.5}} \\
    & PPT~\cite{b45} &ACMMM 25 & 1.1 (4.07\%)  & 11.3\dtplus{$\uparrow$137.4\%} &90.53\dplus{+3.10} &88.81\dplus{+0.69} &84.94\dplus{+1.87}  &93.2{\dplus{+0.5}} \\
    & GFT~\cite{b22} &WACV 26 & 0.7 (4.07\%)  & 6.4\dtplus{$\uparrow$34.45\%} & 89.50\dplus{+2.07} &88.30\dplus{+0.18} &84.63\dplus{+1.56}  &93.0{\dplus{+0.3}} \\

    &PAT(\textbf{ours}) &This paper &\textbf{0.5} (2.26\%)  & \textbf{4.2}\dplus{$\downarrow$11.76\%} & \underline{91.22}\dplus{+3.79} & \textbf{89.67}\dplus{+1.55} &85.25\dplus{+2.18}  & \textbf{93.6}{\dplus{+0.9}} \\

    \hline

    \multirow{9}{*}{\tabincell{c}{ACT~\cite{b11}}} & \textcolor{gray}{Full}  & & \textcolor{gray}{22.1 (100\%)}  & \textcolor{gray}{4.76} & \textcolor{gray}{93.29} & \textcolor{gray}{91.91} & \textcolor{gray}{88.21} & \textcolor{gray}{93.2}\\
    & IDPT~\cite{b19} &ICCV 23  & 1.7 (7.69\%) & 7.1\dtplus{$\uparrow$49.16\%} & 93.12{\dtplus{-0.17}} & \underline{92.26}{\dplus{+0.35}} & 87.65{\dtplus{-0.56}}   & \underline{93.3}\dplus{+0.1}\\
    & DAPT~\cite{b23} &CVPR 24& 1.1 (4.97\%)  & 5.0\dtplus{$\uparrow$5.04\%} &92.60\dtplus{-0.69} & 91.57{\dtplus{-0.34}} & 87.54{\dtplus{-0.67}}  & 92.7{\dtplus{-0.5}}  \\
    & Point-PEFT~\cite{b47} &AAAI 24 &0.7 (3.13\%)  & 7.6\dtplus{$\uparrow$37.37\%} & 90.36{\dtplus{-2.41}} & 90.02{\dtplus{-2.58}} & 85.74{\dtplus{-3.65}} &93.1\dtplus{-0.1} \\

    & PointGST~\cite{b18}  &TPAMI 25&\underline{0.6} (2.77\%)  &\underline{4.8}\dtplus{$\uparrow$0.84\%} &\textbf{93.46}\dplus{+0.17} &\textbf{92.60}\dplus{+0.69} & \textbf{88.27}{\dplus{+0.06}}  &\textbf{93.4}{\dplus{+0.2}} \\

    & PointLoRA~\cite{b25} &CVPR 25 & 0.8 (3.43\%)  &5.1\dtplus{$\uparrow$7.14\%}  &92.60{\dtplus{-0.69}}  &91.39{\dtplus{-0.52}}  &87.89{\dtplus{-0.32}}   &\textbf{93.4}\dplus{+0.2}  \\
    & PPT~\cite{b45} &ACMMM 25 & 1.1 (3.43\%)  &11.3\dtplus{$\uparrow$137.4\%}  &92.77{\dtplus{-0.52}}  &91.57{\dtplus{-0.34}}  &87.78{\dtplus{-0.43}}   &\underline{93.3}\dplus{+0.1}  \\
    &GFT~\cite{b22} &WACV 26 &0.7 (3.43\%)  &6.4\dtplus{$\uparrow$34.45\%}  &92.43{\dtplus{-0.86}}  &91.22{\dtplus{-0.69}}  &86.81{\dtplus{-1.40}}   &93.1\dtplus{-0.1}  \\
    &PAT(\textbf{ours}) &This paper& \textbf{0.5} (2.26\%)  & \textbf{4.2}\dplus{$\downarrow$11.76\%} &\underline{93.29}\dplus{+0.00} & 92.08\dplus{+0.17} & \underline{87.96}{\dtplus{-0.25}}  & \textbf{93.4}{\dplus{+0.2}} \\

\bottomrule
    \end{tabular}
\end{table*}

In the $h$-th encoder layer, two TAMs extract tokens ${\bf{\bar X}}_h^{{\rm{MHA}}} \in {\mathbb{R}^{(m + 1) \times d}}$ ($1 < m < n$) and ${\bf{\bar X}}_h^{{\rm{FFN}}} \in {\mathbb{R}^{(m + 1) \times d}}$ from input tokens $[{\bf{c}}_h^{{\rm{MHA}}};{\bf{Z}}_h^{{\rm{MHA}}}] \in {\mathbb{R}^{(n + 1) \times d}}$ and $[{\bf{c}}_h^{{\rm{FFN}}};{\bf{Z}}_h^{{\rm{FFN}}}] \in {\mathbb{R}^{(n + 1) \times d}}$, respectively. When directly processing $[{\bf{c}}_h^{{\rm{MHA}}};{\bf{Z}}_h^{{\rm{MHA}}}]$ and $[{\bf{c}}_h^{{\rm{FFN}}};{\bf{Z}}_h^{{\rm{FFN}}}]$, the main computational complexities introduced by the MHA and FFN blocks are ${\rm \mathcal{O}}({(n + 1)^2}d)$ and ${\rm \mathcal{O}}((n + 1)qd)$, where $q$ is the hidden dimension of the FFN block. These complexities become ${\rm \mathcal{O}}({(m + 1)^2}d)$ and ${\rm \mathcal{O}}((m + 1)qd)$ when the two blocks process only these extracted tokens. For the MHA and FFN blocks, their computational complexities are reduced by factors of $\frac{{{{(m + 1)}^2}}}{{{{(n + 1)}^2}}}$ and $\frac{{m + 1}}{{n + 1}}$, respectively.

\section{Experiments}

Most 3D PEFT methods are evaluated on object recognition and part segmentation tasks using ModelNet40 \cite{b53}, ScanObjectNN \cite{b54} and ShapeNetPart \cite{b55} datasets. The pre-trained backbones commonly adopted by these methods include Point-MAE \cite{b8}, Point-BERT \cite{b9}, RECON \cite{b10} and ACT \cite{b11}. To ensure fair comparisons, we also evaluate PAT on these datasets with the four backbones.  We compare our method against full fine-tuning (denoted as Full) and representative 3D PEFT approaches including IDPT \cite{b19}, DAPT \cite{b23}, Point-PEFT \cite{b47}, PointGST \cite{b18}, PointLoRA \cite{b25}, PPT \cite{b45} and GFT \cite{b22}. PAT and all comparison methods follow the same data augmentation techniques, optimization settings and data splits used by the corresponding full fine-tuning baselines. Accordingly, for the comparison methods, we directly adopt the results reported in their original papers if they satisfy the above requirements; otherwise, we reproduce the experimental results by running their released code under the same settings. In PAT, we select the rank upper bound $r$ from $\{16, 32, 48\}$, the number of position anchors $m$ from $\{0.6n, 0.8n\}$, and the balance parameter $\eta$ from $\{0.1, 1\}$. In addition, the shared ratio $b$ and the number of neighbors $\mu$ are set to 0.5 and 3, respectively.

\subsection{Experiments on Point Cloud Object Recognition}

The experimental results on the ModelNet40 dataset and the three variants of the ScanObjectNN dataset are reported in TABLE~\ref{tab1}. Based on these results, we have the following observations. 1) As full fine-tuning updates all parameters in these backbones, it incurs a large number of trainable parameters. In contrast, all PEFT methods introduce significantly fewer trainable parameters, while some of them, such as PointGST and PAT, perform comparably to (or even better than) full fine-tuning. 2) Compared to IDPT, DAPT, Point-PEFT, PointLoRA, PPT and GFT, PointGST generally achieves the highest accuracy across all backbones and datasets. According to the Wilcoxon signed-rank test \cite{b56}, PAT performs comparably to PointGST, with no statistically significant difference at the 0.05 significance level. 3) IDPT, DAPT, Point-PEFT, PointGST, PointLoRA, PPT and GFT use a small number of trainable parameters to adapt the pre-trained backbones, but all of them result in additional inference cost. 4) Among 3D PEFT methods, PAT has a clear advantage over the others in terms of parameter count and inference cost. Taking the pre-trained Point-MAE backbone as an example, PAT reduces the number of trainable parameters by 70.6\%, 54.5\%, 28.6\%, 16.7\%, 37.5\%, 54.5\% and 28.6\%, compared to IDPT, DAPT, Point-PEFT, PointGST, PointLoRA, PPT and GFT, respectively. Meanwhile, it reduces GFLOPs by 40.8\%, 16.0\%, 44.7\%, 12.5\%, 17.6\%, 62.8\% and 34.4\%, respectively. Similar phenomena can be observed on the other pre-trained backbones. Overall, these observations demonstrate that PAT achieves performance comparable to state-of-the-art methods, while offering clear advantages in both parameter count and inference overhead.

\begin{table}[!htp]

  \centering
 \footnotesize
  \renewcommand{\arraystretch}{0.94}
  \setlength\tabcolsep{2.0pt}
  \caption{Few-shot classification results on the ModelNet40 dataset. Results are reported as mean accuracy (\%) $\pm$ standard deviation (\%) over 10 independent experiments.}
    \label{tab2}
    \begin{tabular}{lccccc}
    \toprule
   \multirow{2}{*}{Method}&\multirow{2}{*}{Reference} & \multicolumn{2}{c}{5-way} & \multicolumn{2}{c}{10-way} \\
\cmidrule(lr){3-4}\cmidrule(lr){5-6}  &        & 10-shot & 20-shot & 10-shot & 20-shot \\
    \hline
    \multicolumn{6}{c}{\textit{Pre-trained Point-MAE backbone}} \\
        \textcolor{gray}{+ Full \cite{b8}}&\textcolor{gray}{ECCV 22} & \textcolor{gray}{96.3$\pm$2.5} & \textcolor{gray}{97.8$\pm$1.8} & \textcolor{gray}{92.6$\pm$4.1} & \textcolor{gray}{95.0$\pm$3.0}\\
   + IDPT \cite{b19}&   ICCV 23    & \underline{97.3}$\pm$2.1& 97.9$\pm$1.1&92.8$\pm$4.1& 95.4$\pm$2.9\\
   + DAPT \cite{b23} & CVPR 24 & 96.8$\pm$1.8  & 98.0$\pm$1.0 &\underline{93.0}$\pm$3.5 & 95.5$\pm$3.2  \\
   + Point-PEFT \cite{b47}  & AAAI 24 & 95.5$\pm$2.9 & 97.6$\pm$1.7 & 91.7$\pm$4.3 & 94.7$\pm$3.0 \\
   + PointGST \cite{b18}& TPAMI 25 &\textbf{98.0}$\pm$1.8 &\textbf{98.3}$\pm$0.9 &\textbf{93.7}$\pm$4.0 &\underline{95.7}$\pm$2.4 \\
   + PointLoRA \cite{b25}&CVPR 25 &96.6$\pm$2.6  &97.7$\pm$1.3 &92.0$\pm$4.2 &95.2$\pm$3.3\\
   + PPT \cite{b45}&ACMMM 25 &96.4$\pm$2.5  &97.9$\pm$2.6 &92.1$\pm$3.7 &95.2$\pm$3.6\\
   + GFT \cite{b22}&WACV 26 &96.3$\pm$3.1  &\textbf{98.3}$\pm$1.5 &92.1$\pm$5.3 &95.1$\pm$2.9\\
      \rowcolor{gray!20}
   + PAT  &This paper  &96.7$\pm$2.3 &\underline{98.1}$\pm$1.4   &92.6$\pm$4.0  &\textbf{95.8}$\pm$2.3 \\

    \hline

          \multicolumn{6}{c}{\textit{Pre-trained RECON backbone}} \\
       \textcolor{gray}{+ Full \cite{b9}}  &\textcolor{gray}{ICML 23} &\textcolor{gray}{97.3$\pm$1.9} & \textcolor{gray}{98.9$\pm$1.2} & \textcolor{gray}{93.3$\pm$3.9} & \textcolor{gray}{95.8$\pm$3.0} \\
   + IDPT \cite{b19}  & ICCV 23    & 96.8$\pm$2.2& 98.6$\pm$0.8& \underline{92.7}$\pm$3.8& \textbf{95.9}$\pm$3.2\\
   + DAPT \cite{b23} & CVPR 24 &95.6$\pm$3.4 &97.3$\pm$2.0&91.9$\pm$4.9&94.5$\pm$3.2 \\
   + Point-PEFT \cite{b47} & AAAI 24 & 95.4$\pm$2.6 & 97.7$\pm$1.4 & 91.5$\pm$5.0 & 95.2$\pm$3.4 \\
   + PointGST \cite{b18} &TPAMI 25&96.9$\pm$2.2 &\underline{98.7}$\pm$0.9 &\textbf{92.9}$\pm$3.8 &\underline{95.8}$\pm$2.8 \\
   + PointLoRA \cite{b25} &CVPR 25 &96.9$\pm$2.7 &\textbf{98.8}$\pm$1.2 &\underline{92.7}$\pm$4.4 &\underline{95.8}$\pm$2.9\\
   + PPT \cite{b45}&ACMMM 25 &\underline{97.0}$\pm$2.7  &\underline{98.7}$\pm$1.6 &92.2$\pm$5.0 &95.6$\pm$2.9\\
   + GFT \cite{b22}&WACV 26 &96.4$\pm$3.2  &97.7$\pm$1.7 &92.0$\pm$4.3 &94.2$\pm$3.0\\

   \rowcolor{gray!20}
     + PAT  &This paper   &\textbf{97.2}$\pm$2.8  &98.1$\pm$1.0  &92.5$\pm$3.9 &\textbf{95.9}$\pm$3.3\\
     \hline

       \multicolumn{6}{c}{\textit{Pre-trained Point-BERT backbone}} \\
   \textcolor{gray}{+ Full \cite{b10}}  &\textcolor{gray}{CVPR 22} &\textcolor{gray}{94.6$\pm$3.1} & \textcolor{gray}{96.3$\pm$2.7} & \textcolor{gray}{91.0$\pm$5.4} & \textcolor{gray}{92.7$\pm$5.1} \\
   + IDPT \cite{b19} & ICCV 23    & 96.0$\pm$1.7& 97.2$\pm$2.6& 91.9$\pm$4.4& 93.6$\pm$3.5\\
   + DAPT \cite{b23} & CVPR 24 &95.8$\pm$2.1 &97.3$\pm$1.3&\underline{92.2}$\pm$4.3&94.2$\pm$3.4 \\
   + Point-PEFT \cite{b47} & AAAI 24 & 95.4$\pm$3.0 & 97.3$\pm$1.9 & 91.6$\pm$4.5 & 94.5$\pm$3.5 \\
   + PointGST \cite{b18} &TPAMI 25&\textbf{96.5}$\pm$2.4 &\underline{97.9}$\pm$2.0 &\textbf{92.7}$\pm$4.2 &\underline{95.0}$\pm$2.8 \\
   + PointLoRA \cite{b25} &CVPR 25 &95.7$\pm$2.5  &97.8$\pm$1.7  &92.0$\pm$4.1 &94.7$\pm$3.0\\
    + PPT \cite{b45}&ACMMM 25 &96.1$\pm$2.8  &97.4$\pm$2.0 &92.1$\pm$4.0 &94.8$\pm$3.2\\
   + GFT \cite{b22}&WACV 26 &96.0$\pm$2.5  &\textbf{98.1}$\pm$1.3 &\underline{92.2}$\pm$4.5 &94.5$\pm$3.7\\
      \rowcolor{gray!20}
     + PAT  &This paper  &\underline{96.4}$\pm$2.0 &97.6$\pm$1.8  &\underline{92.2}$\pm$3.6  &\textbf{95.2}$\pm$2.9\\
    \hline

              \multicolumn{6}{c}{\textit{Pre-trained ACT backbone}} \\
       \textcolor{gray}{+ Full \cite{b11}}  &\textcolor{gray}{ICLR 23} &\textcolor{gray}{96.8$\pm$2.3} & \textcolor{gray}{98.0$\pm$1.4} & \textcolor{gray}{93.3$\pm$4.0} & \textcolor{gray}{95.6$\pm$2.8} \\
   + IDPT \cite{b19} & ICCV 23    & 96.7$\pm$2.6& 98.2$\pm$0.9& 92.4$\pm$4.5& 95.5$\pm$3.0\\
   + DAPT \cite{b23} & CVPR 24 &94.4$\pm$3.9 &97.3$\pm$2.4&90.6$\pm$6.1&95.1$\pm$3.7 \\
   + Point-PEFT \cite{b47} & AAAI 24 & 95.4$\pm$2.7 & 97.7$\pm$1.7 & 91.8$\pm$5.1 & 94.9$\pm$3.0 \\
   + PointGST \cite{b18} &TPAMI 25&\textbf{97.2}$\pm$1.9 &\underline{98.6}$\pm$1.5 &\textbf{92.8}$\pm$4.0 &95.5$\pm$3.1 \\
   + PointLoRA \cite{b25} &CVPR 25 &\underline{96.9}$\pm$2.7 &\textbf{98.8}$\pm$1.2 &\underline{92.7}$\pm$4.4 &\textbf{95.8}$\pm$2.9\\
   + PPT \cite{b45}&ACMMM 25 &96.8$\pm$2.0  &97.9$\pm$1.3 &92.6$\pm$4.1 &95.4$\pm$3.3\\
   + GFT \cite{b22}&WACV 26 &96.0$\pm$2.3  &98.3$\pm$1.5 &92.1$\pm$5.0 &95.0$\pm$3.4\\

   \rowcolor{gray!20}
     + PAT  &This paper   &\underline{96.9}$\pm$2.4  &98.2$\pm$1.1  &\textbf{92.8}$\pm$4.2 &\underline{95.6}$\pm$3.1\\

    \bottomrule
    \end{tabular}

\end{table}

In line with previous studies \cite{b18,b19,b23,b25}, we evaluate PAT under low-data regimes with few-shot experiments performed on the ModelNet40 dataset. As shown in TABLE~\ref{tab2}, PAT obtains the highest or second-highest accuracy in 10 out of 16 cases across the four pre-trained backbones, in comparison with existing 3D PEFT methods. Although PAT does not achieve top-2 accuracy in some cases, its performance remains competitive. For instance, on the pre-trained RECON backbone with the 10-way 10-shot setting, PAT underperforms IDPT, PointLoRA and PointGST, but it performs better than DAPT, Point-PEFT, PPT and GFT. All results above illustrate the effectiveness of PAT in low-data scenarios.

\begin{table}[htp]

  \centering
 \footnotesize
  \renewcommand{\arraystretch}{0.94}
    \setlength\tabcolsep{0.8pt}
  \caption{Part segmentation performance on the ShapeNetPart dataset. CmIoU (\%) and ImIoU (\%) denote the class-mean and instance-mean mIoU, respectively.}
  \label{tab3}
    \begin{tabular}{lccccc}
    \toprule
    Method & Reference &$\Delta G$ & Params. (M)& CmIoU(\%) & ImIoU(\%) \\
    \hline
       \multicolumn{6}{c}{\textit{Pre-trained Point-MAE backbone}} \\
   \textcolor{gray}{+ Full \cite{b8}}&  \textcolor{gray}{ECCV 22}&- & \textcolor{gray}{27.06} & \textcolor{gray}{84.19} & \textcolor{gray}{86.1} \\
    + IDPT \cite{b19} & ICCV 23 &$\uparrow$& 5.69  & 83.79  & \underline{85.7}  \\
    + DAPT \cite{b23} & CVPR 24 &$\uparrow$ & 5.65  &\underline{84.01} & \underline{85.7} \\
    + Point-PEFT \cite{b47} & AAAI 24 &$\uparrow$& 5.62  & 83.20 &  85.2 \\
    + PointGST \cite{b18} & TPAMI 25 &$\uparrow$ &\underline{5.59}  & 83.81 & \textbf{85.8} \\
    + PointLoRA \cite{b25} &CVPR 25 &$\uparrow$ &5.71 &83.74 &85.3\\
    + PPT \cite{b45}&ACMMM 25 &$\uparrow$ &5.62&\textbf{84.07}&\underline{85.7}\\
       \rowcolor{gray!20}
    + PAT &This paper  &$\downarrow$\textbf{12.8\%} &\textbf{5.55} &83.99 &\textbf{85.8}\\
    \hline

\multicolumn{6}{c}{\textit{Pre-trained RECON backbone}} \\
    \textcolor{gray}{+ Full \cite{b9}} &  \textcolor{gray}{ICML 23} &- & \textcolor{gray}{27.06} & \textcolor{gray}{84.52} & \textcolor{gray}{86.1} \\
    + IDPT \cite{b19} & ICCV 23 &$\uparrow$& 5.69  & 83.66  & \underline{85.7}  \\
    + DAPT \cite{b23} & CVPR 24 &$\uparrow$& 5.65  & 83.87 &\underline{85.7} \\
    + Point-PEFT \cite{b47} & AAAI 24 &$\uparrow$& 5.62  & 83.10  &  85.1 \\
    + PointGST \cite{b18} &TPAMI 25 &$\uparrow$&\underline{5.58}  &\underline{83.98} & \textbf{85.8} \\
    + PointLoRA \cite{b25} &CVPR 25  &$\uparrow$&5.63  &\underline{83.98} &85.4 \\
    + PPT \cite{b45}&ACMMM 25 &$\uparrow$&5.62&\textbf{84.23}&85.6\\
       \rowcolor{gray!20}
    + PAT &This paper &$\downarrow$\textbf{12.8\%}&\textbf{5.55}&\underline{83.98}&\underline{85.7} \\
    \hline

\multicolumn{6}{c}{\textit{Pre-trained Point-BERT backbone}} \\
    \textcolor{gray}{+ Full \cite{b10}}&  \textcolor{gray}{CVPR 22}&- & \textcolor{gray}{27.06} & \textcolor{gray}{84.11} & \textcolor{gray}{85.6} \\
    + IDPT \cite{b19} & ICCV 23 &$\uparrow$& 5.69  & 83.50  & 85.3  \\
    + DAPT \cite{b23} & CVPR 24 &$\uparrow$& 5.65  &83.83 &85.5 \\
    + Point-PEFT \cite{b47} & AAAI 24 &$\uparrow$& 5.62  & 81.12  &  84.3 \\
    + PointGST \cite{b18} &TPAMI 25 &$\uparrow$& \underline{5.58}  & \underline{83.87} & \textbf{85.7} \\
    + PointLoRA \cite{b25} &CVPR 25 &$\uparrow$&5.68 &82.61&85.2\\
    + PPT \cite{b45}&ACMMM 25 &$\uparrow$ &5.62&83.70&85.5\\

       \rowcolor{gray!20}
    + PAT &This paper &$\downarrow$\textbf{12.8\%}&\textbf{5.55}&\textbf{83.92} &\underline{85.6} \\

      \hline

  \multicolumn{6}{c}{\textit{Pre-trained ACT backbone}} \\
    \textcolor{gray}{+ Full \cite{b11}} &  \textcolor{gray}{ICML 23} &-& \textcolor{gray}{27.06} & \textcolor{gray}{84.52} & \textcolor{gray}{86.1} \\
    + IDPT \cite{b19} & ICCV 23 &$\uparrow$& 5.69  & 83.82  &\textbf{85.9}  \\
   + DAPT  \cite{b23} & CVPR 24 &$\uparrow$& 5.65  & 83.44 &85.5 \\
    + Point-PEFT \cite{b47} & AAAI 24 &$\uparrow$& 5.62  & 81.25 &  84.4 \\

    + PointGST \cite{b18} &TPAMI 25 &$\uparrow$&\underline{5.59}  &\textbf{84.04} &  \underline{85.8} \\
    + PointLoRA \cite{b25} &CVPR 25  &$\uparrow$&5.63  &83.13 &85.3 \\
    + PPT \cite{b45}&ACMMM 25 &$\uparrow$&5.62&83.79&85.4\\
       \rowcolor{gray!20}
    + PAT &This paper &$\downarrow$\textbf{12.8\%}&\textbf{5.55}&\underline{84.01}&85.7 \\

    \bottomrule
    \end{tabular}

\end{table}

\subsection{Experiments on Point Cloud Part Segmentation}

We report the part segmentation results on the ShapeNetPart dataset in TABLE~\ref{tab3}, where $\Delta G$ represents the change in GFLOPs relative to the full fine-tuning baseline. The results of GFT are not included in this table, as its part segmentation head differs from those used in the other 3D PEFT methods. Although full fine-tuning outperforms the best 3D PEFT method in most cases, it requires significantly more trainable parameters. In this subsection, we focus on the comparisons among 3D PEFT methods. As illustrated, PAT performs similarly to these competitive PEFT methods while introducing the fewest parameters. Since full fine-tuning merely updates the pre-trained backbones, it brings no extra inference cost. Nevertheless, the trainable modules designed in IDPT, DAPT, Point-PEFT, PointGST, PointLoRA and PPT introduce extra computational overhead at the inference stage. In contrast, PAT has a clear advantage in inference cost. For example, it reduces the GFLOPs of the pre-trained Point-MAE backbone by 12.8\%, compared with full fine-tuning.

\subsection{Ablation Studies}

PAT applies token aggregation-expansion pairs, comprising TAMs and TEMs, to reduce the computational cost of the MHA and FFN blocks. Each TAM incorporates local information into anchor tokens and captures task-specific representations via BSLoRA, which includes a base-sharing scheme and a similarity-regularization term (SRT). In this section, we investigate the effectiveness of local information, base-sharing and SRT in PAT. All ablation experiments are conducted on the PB\_T50\_RS dataset with the pre-trained Point-MAE backbone.

The ablation results are reported in TABLE~\ref{tab4}, where '$\checkmark$' and '$\times$' denote that the corresponding item is selected and unselected, respectively. For clarity, we define the notations in the first column. For example, PAT\_L\&S refers to the PAT variant with both local information and SRT removed; FTOPH is the abbreviation of ``fine-tuning only the prediction head''. The difference between PAT\_L\&B\&S and FTOPH is that PAT\_L\&B\&S learns the representations of anchor tokens for all MHA and FFN blocks while FTOPH considers all input tokens. Based on PAT\_L\&B\&S, PAT\_B\&S further incorporates the local information of anchor tokens. FTOPH exceeds PAT\_L\&B\&S by 1.11\%, but its gain over PAT\_B\&S is negligible (only 0.03\%). This suggests that anchor tokens lead to the loss of some essential information, but the loss can be mitigated by the local information of these tokens. PAT\_L\&B\&S lags behind PAT\_B\&S, PAT\_S and PAT by 1.08\%, 9.23\% and 10.55\%, respectively. A similar trend can also be observed by comparing PAT\_L\&B\&S, PAT\_L\&S, PAT\_L and PAT. These observations indicate that local information, base-sharing and SRT contribute significantly to the performance improvement. According to these results, base-sharing is the most important component in PAT.

\begin{table}[!htb]
  \centering
  \caption{Ablation experiments on the PB\_T50\_RS dataset.}
  \label{tab4}
 \footnotesize
     \renewcommand{\arraystretch}{1.0}
  \setlength\tabcolsep{2.2pt}
  \begin{tabular}{l|ccc|c}

    \toprule
      \textbf{Notation}  &Local Information  &Base-Sharing  &SRT  &\textbf{Accuracy}(\%)\\ \hline

   PAT\_L\&B\&S &$\times$ &$\times$ &$\times$       &74.91 \\

   PAT\_L\&S &$\times$  &$\checkmark$ &$\times$       &80.81  \\
   PAT\_L  &$\times$     &$\checkmark$    &$\checkmark$    	&82.89\\

   PAT\_B\&S    &$\checkmark$  &$\times$  &$\times$ &75.99    \\
   PAT\_S    &$\checkmark$  &$\checkmark$  &$\times$ &84.14    \\

        \rowcolor{gray!20}
    PAT &$\checkmark$    &$\checkmark$ &$\checkmark$ &\textbf{85.46}\\ \hline

    FTOPH&\multicolumn{3}{c|}{\textit{Fine-tuning only the prediction head}} &76.02\\

\bottomrule
  \end{tabular}
\end{table}

\begin{figure}[!htb]
\centering
\includegraphics[trim={0.1cm 0.50cm 0.0cm 0.36cm},clip,width=3.4in]{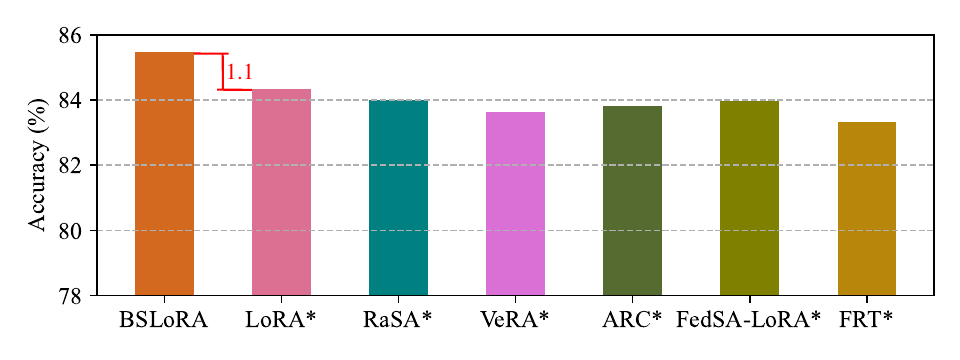}
\caption{Performance verification of BSLoRA.}
\label{fig4}
\end{figure}

Here, we validate the effectiveness of BSLoRA by comparing it with full-rank tuning (FRT), LoRA and representative LoRA-based methods that incorporate shared components among LRMs (such as RaSA \cite{b41}, VeRA \cite{b42}, ARC \cite{b43} and FedSA-LoRA \cite{b44}). In our experiments, we iteratively replace BSLoRA with FRT, LoRA, RaSA, VeRA, ARC and FedSA-LoRA. The resulting variants are denoted as FRT*, LoRA*, RaSA*, VeRA*, ARC* and FedSA-LoRA*, respectively. Among them, BSLoRA, LoRA*, RaSA*, VeRA*, ARC* and FedSA-LoRA* are PEFT methods. This means that they introduce considerably fewer trainable parameters than FRT*. We display the comparison results in Fig.~\ref{fig4}. BSLoRA's result is obtained with $b=0.5$ and $r=32$. Although FRT* introduces the largest number of trainable parameters, its performance may suffer from overfitting, which explains why it performs worse than the others. BSLoRA exceeds the best method among RaSA*, VeRA*, ARC* and FedSA-LoRA* by 1.5\%. All these experimental results demonstrate the effectiveness of BSLoRA. Compared with LoRA, BSLoRA reduces the number of trainable parameters by 47.9\% (see Fig.~\ref{fig3}) while outperforming it by 1.1\% in terms of accuracy. Moreover, the ranks of all low-rank matrices learned by both BSLoRA and LoRA are 32, demonstrating that BSLoRA maintains the same rank upper bound as LoRA.

\subsection{Further Exploration}

The performance of PAT has been effectively verified on object recognition and part segmentation tasks with widely used pre-trained backbones. To further evaluate the scalability of PAT, we extend it to semantic segmentation and object detection tasks and incorporate it into the PointGPT-L backbone \cite{b57}, which has a considerably larger scale than the four backbones mentioned above. Given that PAT is designed for the efficient adaptation of pre-trained point cloud transformers, we analyze its inference speed and GPU consumption. Additionally, we compare PAT with existing PEFT methods in NLP and 2D vision.

\noindent\textit{1) Experiments on the Pre-Trained PointGPT-L Backbone}

In \cite{b57}, the transfer performance of the pre-trained PointGPT-L backbone is evaluated on the ScanObjectNN (including OBJ\_BG, OBJ\_ONLY and PB\_T50\_RS), ModelNet40 and ShapeNetPart datasets. By incorporating PAT into the pre-trained backbone, we also conduct related experiments on these benchmarks. Following the original work \cite{b57}, we set the learning rate to 0.0001 and the number of training epochs to 50 for object recognition tasks. In PAT, the balance parameter $\eta$ is fixed at 1. Since the transfer abilities of IDPT, DAPT, Point-PEFT and PointGST have been verified on the pre-trained PointGPT-L backbone, we select them as comparison methods.

\begin{table*}[tp]

  \centering
  \renewcommand{\arraystretch}{1.0}
  \setlength\tabcolsep{3.1pt}
\footnotesize
  \caption{Classification accuracy (\%) of different fine-tuning methods using the pre-trained PointGPT-L backbone. ``$x$w$y$s'' denotes the $x$-way $y$-shot setting.}
    \label{tab5}
    \begin{tabular}{lcccccccccc}
    \toprule
     \multirow{2}{*}{Method}  &\multicolumn{6}{c}{Shape Classification} &\multicolumn{4}{c}{Few-Shot Learning}  \\
    \cmidrule(lr){2-7} \cmidrule(lr){8-11}  &\#Param. (M) &\#GFLOPs &OBJ\_BG & OBJ\_ONLY &PB\_T50\_RS &ModelNet40   &5w10s&5w20s&10w10s&10w20s \\
    \hline
   \textcolor{gray}{Full}   & \textcolor{gray}{360.5 (100\%)}  & \textcolor{gray}{67.71(-)} & \textcolor{gray}{97.2} & \textcolor{gray}{96.6} & \textcolor{gray}{93.4} & \textcolor{gray}{94.1} & \textcolor{gray}{98.0$\pm$1.9} & \textcolor{gray}{99.0$\pm$1.0} & \textcolor{gray}{91.1$\pm$3.3} & \textcolor{gray}{96.1$\pm$2.8} \\
     IDPT \cite{b19}  & 10.0 (2.77\%) & 75.19\dtplus{$\uparrow$11.05\%} &98.11\dplus{+0.91} &96.04\dtplus{-0.56} & 92.99{\dtplus{-0.41}}  &93.4\dtplus{-0.7} & 96.8$\pm$2.0&98.3$\pm$1.4&95.3$\pm$2.5& 96.5$\pm$2.5\\
     DAPT \cite{b23}& 4.2 (1.17\%)  & 71.64\dtplus{$\uparrow$5.80\%} & 98.11\dplus{+0.91} &96.21\dtplus{+0.39} & 93.02{\dtplus{-0.38}}  & 94.2{\dplus{+0.1}} &97.2$\pm$3.4 & 98.4$\pm$1.3& 95.9$\pm$2.7& 97.3$\pm$2.4\\
     Point-PEFT \cite{b47} &3.1 (0.86\%)  & 73.62\dtplus{$\uparrow$8.73\%} & 96.39{\dtplus{-0.81}} & 94.66\dtplus{-1.94} & 92.85{\dtplus{-0.55}} &93.5\dtplus{-0.6} &95.7$\pm$2.5&98.0$\pm$1.8&94.8$\pm$2.7& 96.3$\pm$2.4\\

     PointGST \cite{b18} &\underline{2.4} (0.67\%) &\underline{67.98}\dtplus{$\uparrow$0.40\%} &\underline{98.97}\dplus{+1.77} &\textbf{97.59}\dplus{+0.99} &\textbf{94.83}\dplus{+1.43}  & \underline{94.8}{\dplus{+0.7}} & 97.4$\pm$2.0 & \textbf{99.6}$\pm$0.5 & \textbf{96.3}$\pm$3.6 & \textbf{97.6}$\pm$2.9\\

    PAT(\textbf{ours}) & \textbf{2.1} (0.58\%) & \textbf{60.60}\dplus{$\downarrow$10.50\%} &\textbf{99.14}\dplus{+1.94} &\underline{97.07}\dplus{+0.47} &\underline{94.76}\dplus{+1.36}  &\textbf{94.9}{\dplus{+0.8}} & \textbf{98.1}$\pm$2.8 &\underline{99.1}$\pm$0.7 & \underline{96.1}$\pm$2.2 & \underline{97.5}$\pm$2.7\\

\bottomrule
    \end{tabular}
\end{table*}

We report the experimental results on the ScanObjectNN and ModelNet40 datasets in TABLE~\ref{tab5}. IDPT, DAPT, Point-PEFT and PointGST require substantially fewer trainable parameters than full fine-tuning, but they incur additional inference overhead. PointGST performs better than IDPT, DAPT and Point-PEFT in terms of classification accuracy, parameter count and inference overhead. Compared with PointGST, PAT achieves comparable classification accuracy while reducing the parameter count and inference GFLOPs by 12.5\% and 10.9\%, respectively. Notably, PAT is the only method that consistently achieves higher accuracy than full fine-tuning in all cases and has advantages in both parameter count and inference overhead. TABLE~\ref{tab6} shows the part segmentation results on the ShapeNetPart dataset. As demonstrated, PAT performs comparably to the best method among IDPT, DAPT, Point-PEFT and PointGST, yet requires fewer GFLOPs and trainable parameters.

\begin{table}[htp]

  \centering
 \footnotesize
  \renewcommand{\arraystretch}{1.0}
    \setlength\tabcolsep{2.8pt}
  \caption{Part segmentation results of different fine-tuning methods using the pre-trained PointGPT-L backbone.}
  \label{tab6}
    \begin{tabular}{lccccc}
    \toprule
    Method  &${\Delta G}$ & Params. (M)& CmIoU(\%) & ImIoU(\%) \\
    \hline
    \textcolor{gray}{Full}  &-& \textcolor{gray}{339.39} & \textcolor{gray}{84.80} & \textcolor{gray}{86.6} \\
    IDPT \cite{b19}&$\uparrow$& 33.17  &82.91  &85.1  \\
   DAPT \cite{b23} &$\uparrow$& 33.85  & 83.03 &85.2 \\
    Point-PEFT \cite{b47} &$\uparrow$& 31.94 & 82.34 &  84.9 \\

    PointGST \cite{b18} &$\uparrow$&\underline{31.85}  &\underline{83.91} &\textbf{85.8} \\

       \rowcolor{gray!20}
    PAT &$\downarrow$\textbf{12.0\%}&\textbf{31.56}&\textbf{83.96}&\underline{85.6} \\

    \bottomrule
    \end{tabular}

\end{table}

\noindent\textit{2) Experiments on Scene-Level Semantic Segmentation and Object Detection}

Among the backbones adopted above, ACT \cite{b11} has demonstrated strong performance on semantic segmentation and object detection tasks using S3DIS \cite{b58} and ScanNetV2 \cite{b59} datasets, respectively. Here, taking the pre-trained ACT backbone as an example, we test the performance of PAT on the two datasets. For fair comparisons, we configure PAT with the same optimization settings and data splits as in PointGST, where the optimization settings include initial learning rate, weight decay, training epochs and batch size. Since PointGST has achieved state-of-the-art results on the datasets among existing 3D PEFT methods \cite{b18}, we select it as the comparison method. Moreover, full fine-tuning is taken as the baseline.

\begin{figure}[!htb]
\centering
\includegraphics[trim={0.1cm 0.50cm 0.0cm 0.36cm},clip,width=3.4in]{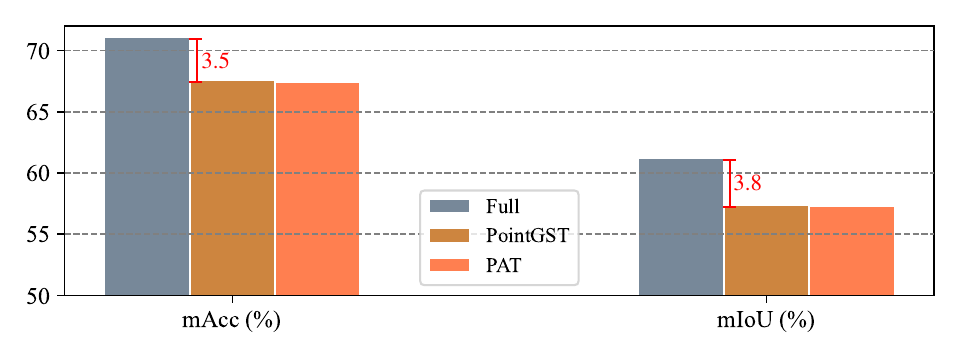}
\caption{Comparisons of mAcc and mIoU on the S3DIS dataset.}
\label{fig5}
\end{figure}

\begin{figure}[!htb]
\centering
\includegraphics[trim={0.1cm 0.50cm 0.0cm 0.36cm},clip,width=3.4in]{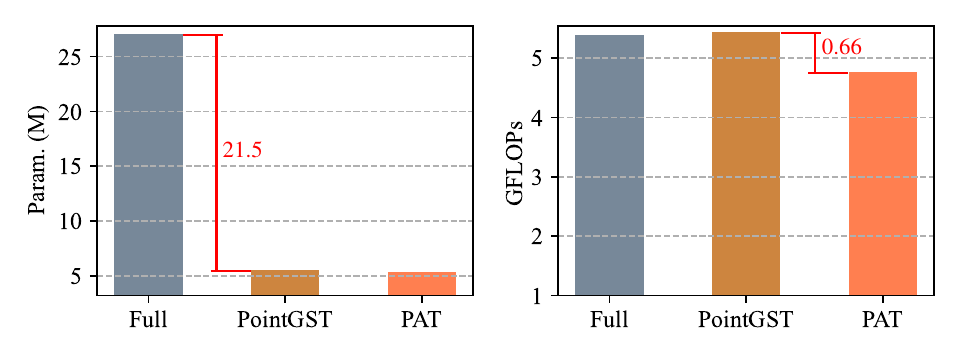}
\caption{Comparisons of the number of trainable parameters and the inference GFLOPs of the backbone on the S3DIS dataset.}
\label{fig6}
\end{figure}

\begin{figure}[!htb]
\centering
\includegraphics[trim={0.1cm 0.50cm 0.0cm 0.36cm},clip,width=3.4in]{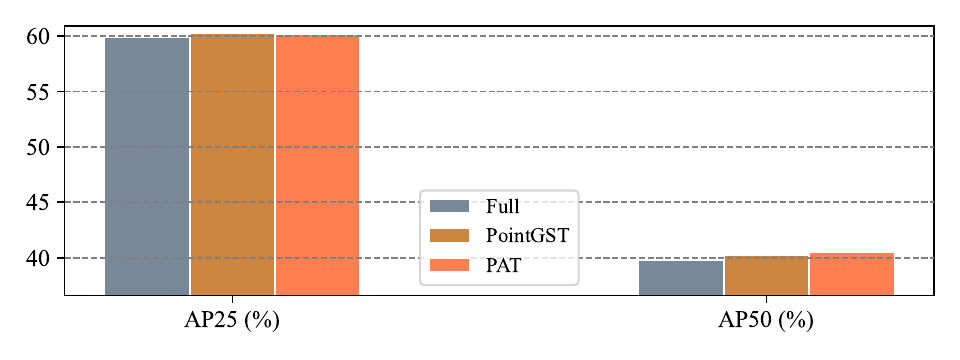}
\caption{Comparisons of AP25 and AP50 on the ScanNetV2 dataset.}
\label{fig7}
\end{figure}

\begin{figure}[!htb]
\centering
\includegraphics[trim={0.1cm 0.50cm 0.0cm 0.36cm},clip,width=3.4in]{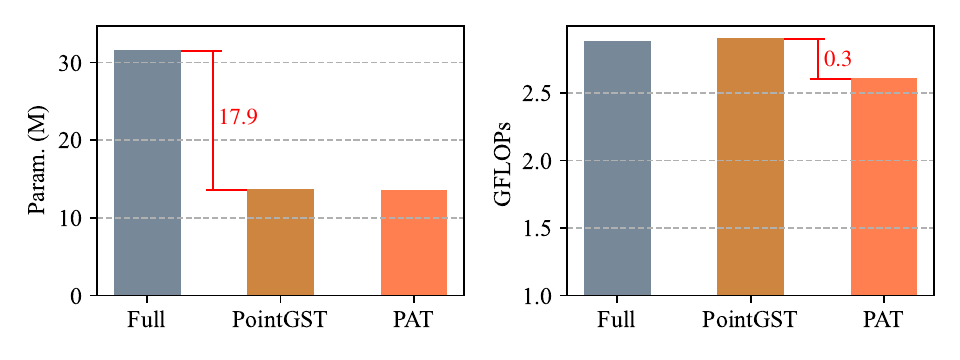}
\caption{Comparisons of the number of trainable parameters and the inference GFLOPs of the backbone on the ScanNetV2 dataset.}
\label{fig8}
\end{figure}

Figs.~\ref{fig5}-\ref{fig6} show the semantic segmentation results. Although full fine-tuning outperforms both PointGST and PAT in terms of mean accuracy (mAcc) and mean IoU (mIoU), it requires updating substantially more parameters, which is inefficient for the adaptation process. Compared to full fine-tuning, PointGST cuts the number of trainable parameters by 79.5\%, at the cost of slightly increasing the GFLOPs of the pre-trained backbone. In contrast, PAT reduces the inference cost while achieving performance similar to PointGST. Specifically, it reduces the GFLOPs by 12.2\% and maintains a similar parameter count, compared to PointGST. Consistent with PointGST, we adopt 3DETR \cite{b60}, a standard detector, as our framework, and replace its default encoder with the pre-trained ACT backbone. The comparisons of object detection results are shown in Figs.~\ref{fig7}-\ref{fig8}, where the detection performance is measured by the average precision at 0.25 and 0.5 IoU thresholds (AP25 and AP50). According to these results, we observe that 1) full fine-tuning, PointGST and PAT perform similarly in terms of AP25 and AP50; 2) PointGST and PAT introduce a comparable number of trainable parameters, and both of them are much more parameter-efficient than full fine-tuning; and 3) compared with full fine-tuning, PAT reduces the GFLOPs by 9.7\%, but PointGST introduces extra inference overhead.

\noindent \textit{3) Inference Speed and GPU Memory Comparisons}

Existing 3D PEFT methods normally introduce additional computational modules during inference, leading to slower inference than full fine-tuning. In contrast, PAT aims to lower the computational cost at the inference stage. Following PointGST, we also report the inference speed (measured in frames per second) and GPU memory usage (with a batch size of 8) of PAT on the pre-trained PointGPT-L backbone using the PB\_T50\_RS dataset. As IDPT, DAPT, Point-PEFT and PointGST have excellent performance on the pre-trained backbone, they are chosen as comparison methods. All experiments in this subsection are carried out on a single RTX 3090 GPU.

\begin{figure}[!htb]
\centering
\includegraphics[trim={0.1cm 0.50cm 0.0cm 0.36cm},clip,width=3.4in]{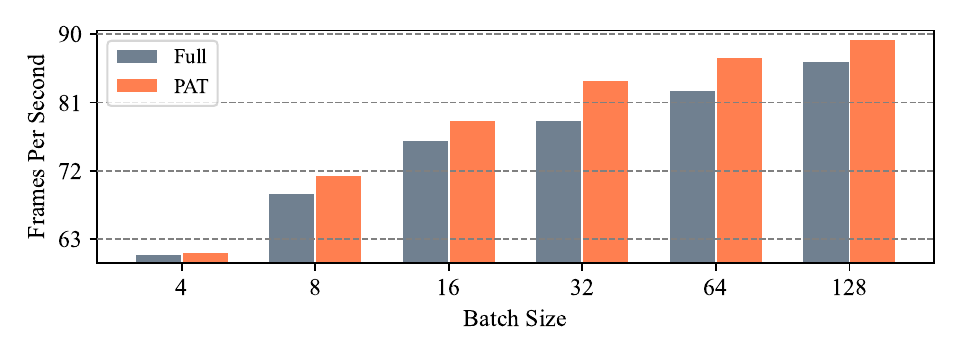}
\caption{Inference speed comparisons with varying batch sizes.}
\label{fig9}
\end{figure}

\begin{figure}[!htb]
\centering
\includegraphics[trim={0.1cm 0.50cm 0.0cm 0.36cm},clip,width=3.4in]{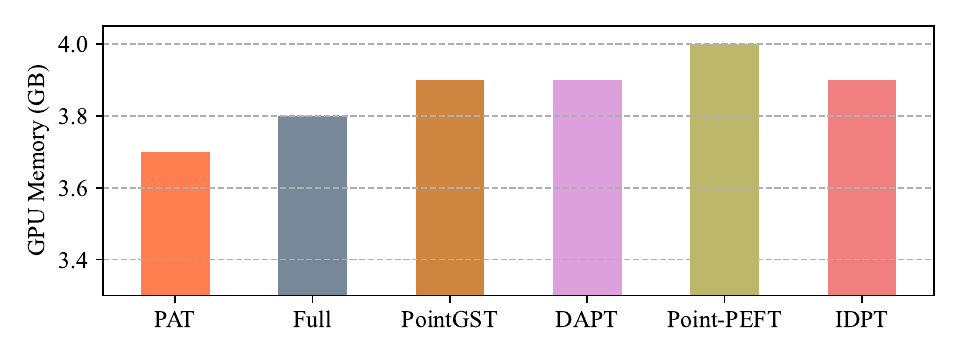}
\caption{Comparisons of GPU memory at a batch size of 8.}
\label{fig10}
\end{figure}

Since existing 3D PEFT methods suffer from slower inference than full fine-tuning, we focus on comparing the inference speed of full fine-tuning and PAT. In Fig.~\ref{fig9}, we display the inference speed with varying batch sizes. Obviously, inference speed is highly influenced by batch size. PAT consistently runs faster than full fine-tuning across all batch sizes. The comparisons in terms of GPU memory (GB) are shown in Fig.~\ref{fig10}. IDPT, DAPT, Point-PEFT and PointGST require more GPU memory than full fine-tuning. However, PAT requires the least GPU memory.

\noindent \textit{4) Comparisons with PEFT Methods in NLP and 2D Vision}

PEFT methods have been extensively studied in both NLP and 2D vision. Here, we further compare PAT with representative PEFT methods from both domains. Consistent with DAPT, PointGST and PointLoRA, we select the pre-trained Point-MAE as the backbone and conduct experiments on the PB\_T50\_RS dataset. TABLE~\ref{tab7} reports the experimental results of different PEFT methods. Although introducing only a small number of trainable parameters, the best method among those designed for NLP and 2D vision still lags behind full fine-tuning by a clear margin. In contrast, each of these 3D PEFT methods can achieve performance similar to that of full fine-tuning. These phenomena demonstrate that PEFT methods from NLP and 2D vision struggle to process 3D point cloud data. Among 3D PEFT methods, PAT achieves the second-highest accuracy, with a gap of only 0.07\% to the best-performing method (PointLoRA). However, compared with PointLoRA, PAT has a clear advantage in computational overhead and parameter count.

\begin{table}[htp]

  \centering
 \footnotesize
  \renewcommand{\arraystretch}{1.0}
    \setlength\tabcolsep{2.8pt}
  \caption{Comparisons with representative PEFT methods designed for NLP and 2D vision on the PB\_T50\_RS dataset. }
  \label{tab7}
    \begin{tabular}{lccccc}
    \toprule
    Method & Design for &${\Delta G}$ & Params. (M)&Accuracy(\%) \\
    \hline
    \textcolor{gray}{Full} & &-& \textcolor{gray}{22.1} & \textcolor{gray}{85.18}  \\
    \hline
    Adapter~\cite{b16} &NLP&$\uparrow$& 0.9  &83.93  \\
    Prefix tuning~\cite{b27} &NLP&$\uparrow$& 0.7  &77.72  \\
    LoRA~\cite{b12} &NLP&-& 0.9  &81.74  \\
    DEPT~\cite{b15} &NLP&$\uparrow$ & 0.3  &79.70  \\
  FourierFT~\cite{b61}&NLP&-& 0.3  &78.57  \\
  LOFT~\cite{b13}&NLP&-&0.9  &83.10  \\
   \hline
    VPT-Deep~\cite{b14}&2D&$\uparrow$ & 0.4  &81.09  \\
    SSF~\cite{b36}&2D&-& 0.4  &82.58  \\
    FacT~\cite{b37}&2D&-& 0.5  &78.76  \\
    SCT~\cite{b62}&2D&$\uparrow$ & 0.3  &80.40  \\
    PEGO~\cite{b38}&2D&-&1.0  &83.27  \\
    DyT~\cite{b63}&2D &$\downarrow$\underline{3.5\%} &1.3  &83.73  \\
     \hline
  IDPT~\cite{b19} &3D&$\uparrow$& 1.7   &84.94  \\
   DAPT~\cite{b23} &3D &$\uparrow$&1.1  & 85.08  \\
    Point-PEFT~\cite{b47} &3D &$\uparrow$&0.7 & 84.94  \\

    PointGST~\cite{b18} &3D &$\uparrow$&0.6  &85.29  \\

    PointLoRA~\cite{b25} &3D &$\uparrow$ &0.8  &\textbf{85.53}  \\
     PPT~\cite{b45} &3D &$\uparrow$&1.1  &85.01  \\
     GFT~\cite{b22} &3D &$\uparrow$&0.7  &85.05  \\
       \rowcolor{gray!20}
    PAT &3D&$\downarrow$\textbf{12.0\%}&0.6&\underline{85.46} \\

    \bottomrule
    \end{tabular}

\end{table}

\section{Conclusion}

In this paper, we propose a novel PEFT method named position anchor tuning (PAT) to adapt pre-trained point cloud transformers to various downstream tasks. Since MHA and FFN are computation-heavy blocks in pre-trained backbones, PAT applies token aggregation-expansion pairs, which consist of token aggregation modules (TAMs) and token expansion modules (TEMs), to save the computational cost of these blocks, thus enhancing inference efficiency. TAMs introduce trainable parameters to capture task-specific representations for downstream tasks and simultaneously extract representative tokens that are processed by MHA and FFN blocks. In contrast, TEMs propagate the extracted representations back to the full set of original tokens, without introducing any trainable parameters. We further propose base-sharing low-rank adaptation (BSLoRA) for TAMs to effectively learn task-specific representations with high parameter efficiency. Extensive experiments are conducted to evaluate the performance of PAT on popular datasets. These experimental results illustrate that PAT achieves performance comparable to state-of-the-art methods while requiring substantially fewer GFLOPs.



\end{document}